\documentclass[10pt,twocolumn,letterpaper]{article}

\usepackage[pagenumbers]{cvpr} 

\usepackage{xcolor,colortbl} 
\usepackage{booktabs,multirow,amssymb}
\definecolor{googlered}{HTML}{DB4437} 
\definecolor{googleblue}{HTML}{4285F4} 
\definecolor{googlegreen}{HTML}{0F9D58} 
\definecolor{googleyellow}{HTML}{F4B400} 
\definecolor{googlepurple}{HTML}{9C27B0} 
\definecolor{googleorange}{HTML}{F4A11B} 
\definecolor{teal}{HTML}{457B9D}
\definecolor{lightblue}{rgb}{0.85,0.92,1.0}

\newcommand{\blueText}[1]{{\color{googleblue}#1}}

\newcommand{\yellowText}[1]{{\color{googleyellow}#1}}

\newcommand{\equalcontrib}{\thanks{Equal contribution.}}
\newcommand{\projectlead}{\thanks{Project lead.}}

 \usepackage{soul}

\DeclareMathOperator*{\argmax}{arg\,max}

\definecolor{cvprblue}{rgb}{0.21,0.49,0.74}
\usepackage[pagebackref,breaklinks,colorlinks,allcolors=cvprblue]{hyperref}

\def\paperID{8527} 
\def\confName{CVPR}
\def\confYear{2026}

\title{VA-$\boldsymbol{\pi}$: Variational Policy Alignment for Pixel-Aware Autoregressive Generation}

\author{%
Xinyao Liao$^{1,2}$\equalcontrib \quad Qiyuan He$^{2}$\footnotemark[1] \projectlead \quad Kai Xu$^{2}$ \quad Xiaoye Qu$^{1}$ 
\quad Yicong Li$^{2}$ \quad Wei Wei$^{1}$ \quad Angela Yao$^{2}$ \\
$^1$Huazhong University of Science \& Technology \quad $^2$National University of Singapore \\
}

\begin{document}
\maketitle
\begin{abstract}

Autoregressive (AR) visual generation relies on tokenizers to map images to and from discrete sequences.  However, tokenizers are trained to reconstruct clean images from ground-truth tokens, while AR generators are optimized only for token likelihood. This misalignment leads to generated token sequences that may decode into low-quality images, without direct supervision from the pixel space. We propose \textbf{VA}-$\boldsymbol{\pi}$, a lightweight post-training framework that directly optimizes AR models with a principled pixel-space objective. VA-$\pi$ formulates the generator–tokenizer alignment as a variational optimization, deriving an evidence lower bound (ELBO) that unifies pixel reconstruction and autoregressive modeling. To optimize under the discrete token space, VA-$\pi$ introduces a reinforcement-based alignment strategy that treats the AR generator as a policy, uses pixel-space reconstruction quality as its intrinsic reward. The reward is measured by how well the predicted token sequences can reconstruct the original image under teacher forcing, giving the model direct pixel-level guidance without expensive free-running sampling.
The regularization term of the ELBO serves as a natural regularizer, maintaining distributional consistency of tokens. 
VA-$\pi$ enables rapid adaptation of existing AR generators, without neither tokenizer retraining nor external reward models. With only 1\% ImageNet-1K data and 25 minutes of tuning, it reduces FID from 14.36 to 7.65 and improves IS from 86.55 to 116.70 on LlamaGen-XXL, while also yielding notable gains in the text-to-image task on GenEval for both visual generation model (LlamaGen: from 0.306 to 0.339) and unified multi-modal model (Janus-Pro: from 0.725 to 0.744). Code is available at {\hypersetup{urlcolor=magenta}\url{https://github.com/Lil-Shake/VA-Pi}}.

\end{abstract}    
\section{Introduction}
\begin{figure}
    \centering
    \begin{minipage}[b]{.9\linewidth}
        \centering
        \includegraphics[width=\linewidth]{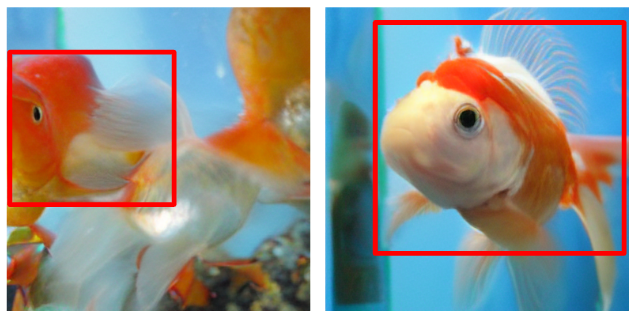}
        \small (a) Qualitative comparison between LlamaGen-XXL (left) and VA-$\pi$ (right) on gold fish image generation
    \end{minipage} \\
    \centering
    \begin{minipage}[b]{0.48\linewidth}
        \centering
        \includegraphics[width=\linewidth]{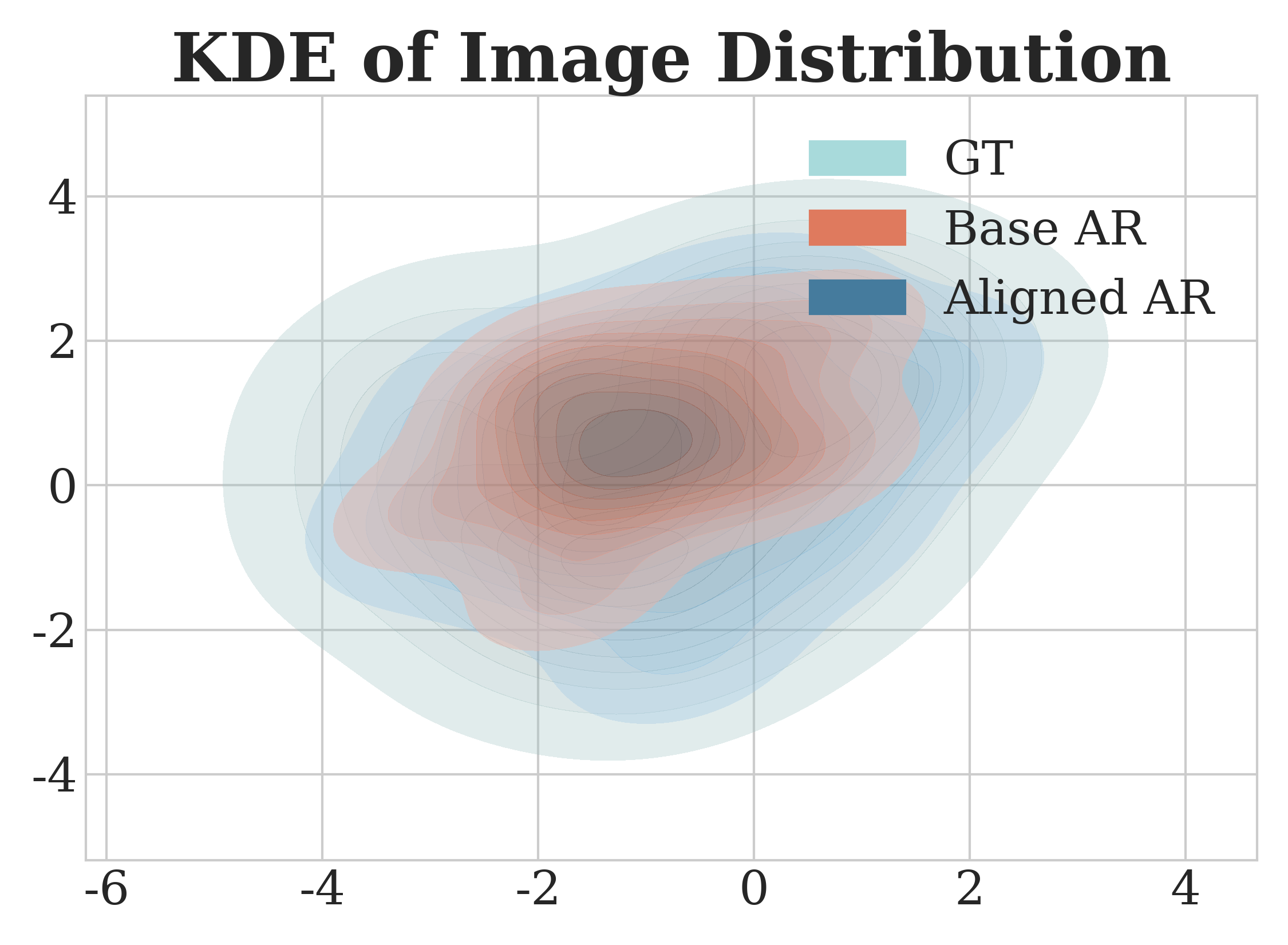}
        \small (b) Kernel Density Estimation (KDE) of image embeddings showing VA-$\pi$ shifts the AR generator’s output closer to the ground-truth manifold.
    \end{minipage}
    \hfill
    \begin{minipage}[b]{0.48\linewidth}
        \centering
        \includegraphics[width=\linewidth]{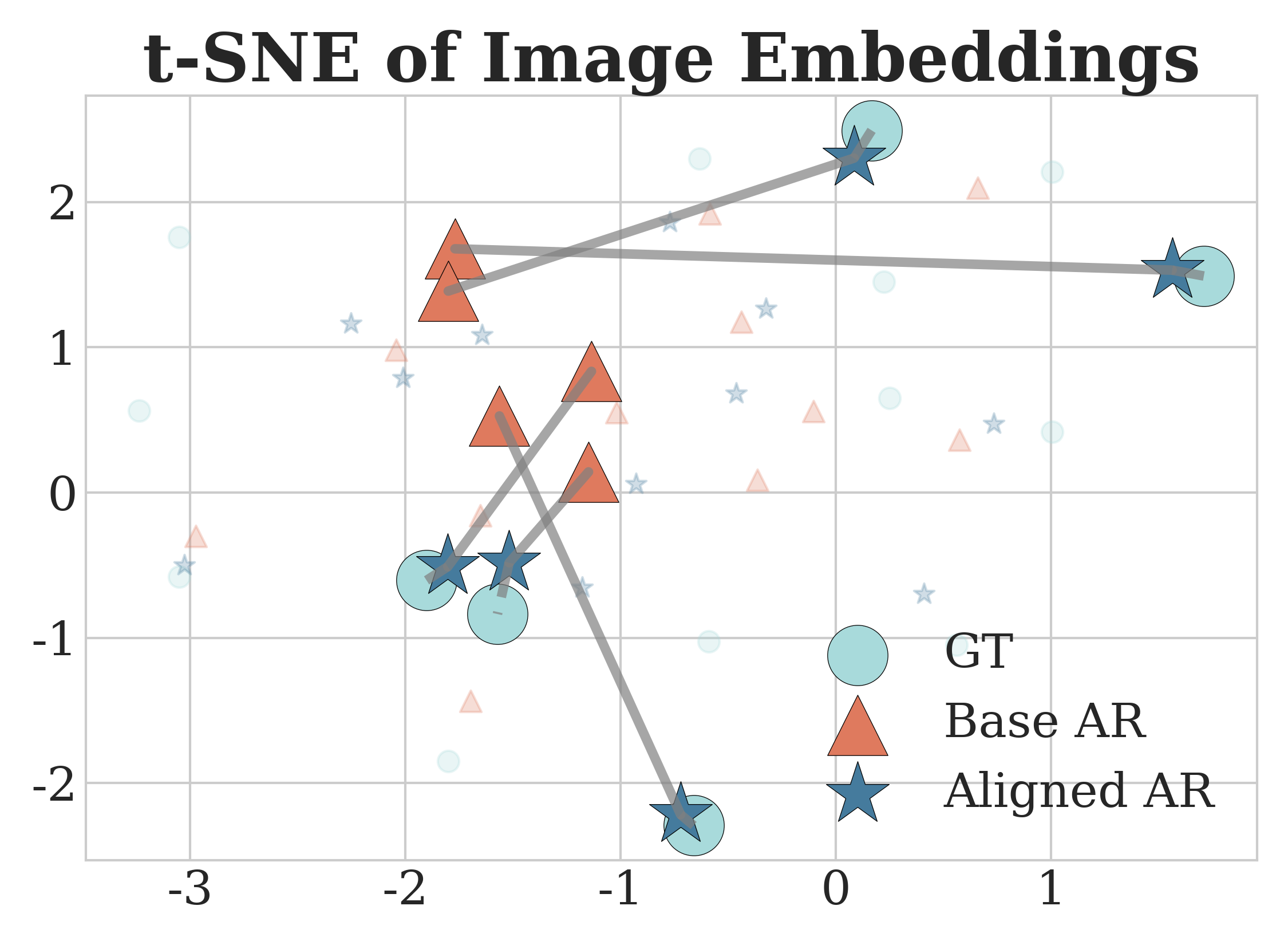}
        \small (c) t-SNE visualization of image embeddings illustrating how VA-$\pi$ aligns the generator’s representation with the ground-truth manifold.
    \end{minipage}
    \caption{\textbf{Pixel-Aware Alignment via VA-$\boldsymbol{\pi}$.} VA-$\pi$ enables efficient post-training via variational policy optimization, aligning the pixel-space distribution of AR generated images with that of ground-truth images.}
    \label{fig:teaser}
\vspace{-10pt}
\end{figure}

\begin{figure*}[t]
    \centering
    \includegraphics[width=.85\textwidth]{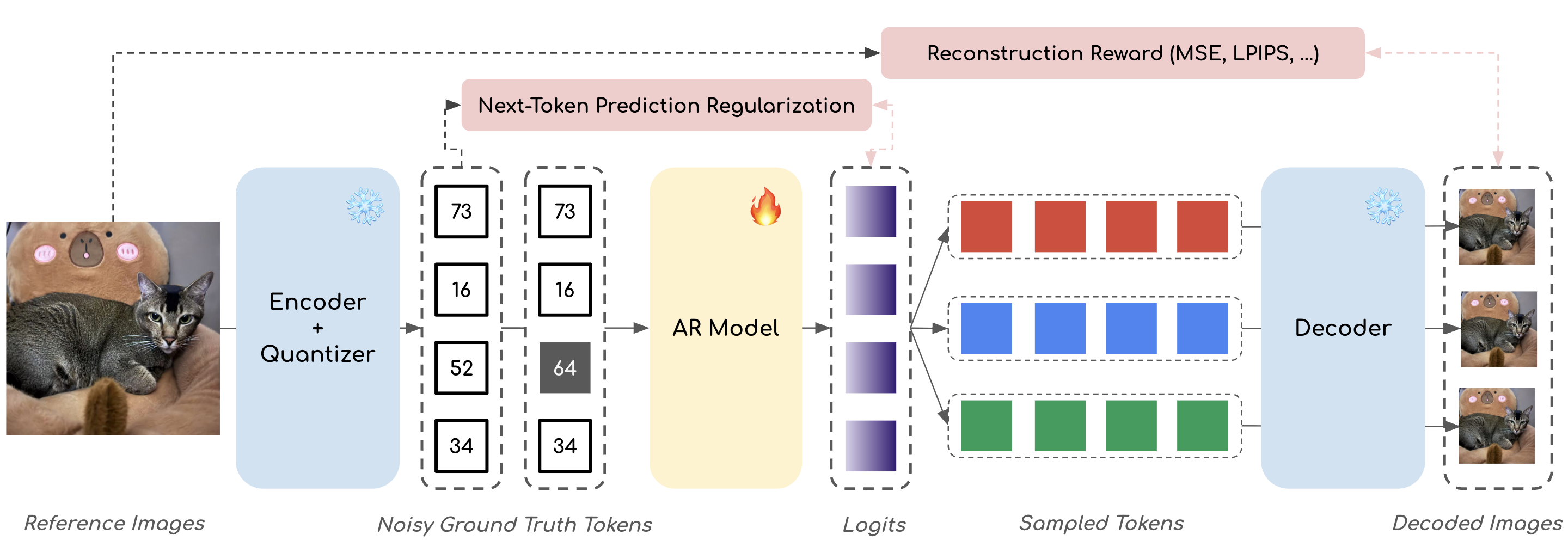}
    \caption{\textbf{Overview of \textbf{VA}-$\boldsymbol{\pi}$}. \textbf{VA}-$\boldsymbol{\pi}$ aligns the visual \yellowText{AR model} with \blueText{tokenizer} via variational optimization. Given a reference image and its ground-truth tokens, VA-$\pi$ adds context noise and lets the AR model compute logits under teacher forcing and samples target tokens. These sampled tokens are decoded back into an image, and the reconstruction reward is defined against the reference image. This reward is then used for policy updates within an RL framework such as GRPO~\citep{yuan2025argrpo}. Additionally, a likelihood regularization using cross-entropy loss between the logits and ground-truth tokens is retained to preserve the model’s original next-token prediction ability.}
    \label{fig:method}
\vspace{-5pt}
\end{figure*}

Autoregressive (AR) image generation models represent visual data as sequences of discrete tokens. This formulation naturally aligns with the architecture of large language models (LLMs)~\citep{gemini, gpt} and paves a way for unified 
multimodal systems~\citep{bai2024sequential, team2024chameleon, chung2024scaling, deng2025bagel, unitok, egotwin}. 
The ultimate goal of any image generation model is to capture the distribution of images in pixel space.
From this perspective, a more principled formulation would be to optimize an end-to-end objective defined directly over the pixel distribution.
But such a direct approach is known to be intractable in practice~\citep{vae, adm}, motivating the use of visual tokenizers to make AR modeling feasible.
The standard pipeline features two stages. 
In the first stage, a visual tokenizer~\cite{vqgan} is trained, with an encoder that converts images into discrete token sequences, and a decoder that reconstructs the images from those tokens. 
In the second stage, an AR model is trained to capture the distribution over these discrete sequences.

However, optimized only at the token level without pixel-space supervision, the AR generator can produce high-likelihood token sequences that decode into visually suboptimal images with artifacts and degraded perceptual quality~\citep{he2025rear, qiu2025robustok}. We refer to these as \textit{off-manifold token sequences}, which, when decoded, deviate from the image manifold and thus produce incoherent visual structures.

Previous studies address the mismatch between token-level likelihood and image-level fidelity by applying noisy-context regularization to either the AR generator~\citep{he2025rear, randar, rar} or the tokenizer~\citep{qiu2025robustok, rae}.
These methods inject noise during training to improve robustness against corrupted sequences. However, they did not directly address the pixel-level misalignment.
Moreover, the tokenizer-centric approach only enables the tokenizer to be more tolerant of off-manifold token sequences rather than preventing their generation from the AR generator.
As shown in Sec.~\ref{sec:main_results_c2i}, training on excessive noise can even overly smooth the decoder’s token-pixel mapping, leading to reduced reconstruction sharpness and visual fidelity.

In this work, we take a standpoint of solving the root cause by directly aligning AR-generated token sequences to the distribution of images in pixel space, fundamentally mitigating the generation of off-manifold token sequences. 
We pose the question as: \textbf{\textit{Can we design an objective that aligns token-level modeling with pixel-level distributions?}} 
We show that this is possible by 
framing the AR-generated discrete token sequence as a latent random variable of the pixel-level image. 
This perspective leads to a tractable evidence lower bound (ELBO) of the image likelihood. 
Specifically, we interpret the pixel reconstruction produced by the tokenizer’s decoder as corresponding to the reconstruction term of the ELBO, while the AR model’s likelihood objective serves as the prior term that preserves proper token-level likelihood modeling.
Such a framing unifies pixel-space reconstruction with token-level predictions.

However, as the variables are discrete, maximizing the ELBO is non-trivial.
A common solution is the straight-through estimator (STE)~\citep{vqvae}, which provides a surrogate gradient path from discrete tokens to continuous logits, enabling gradient flow into the AR generator.
Nevertheless, STE only propagates gradients along the ground-truth path, limiting learning to observed token sequences.

To overcome this, we propose \underline{\textbf{V}}ariational Policy \underline{\textbf{A}}lignment for \underline{\textbf{Pi}}xel-aware Autoregressive Generation, or \textbf{VA}-$\boldsymbol{\pi}$ for short. 
VA-$\pi$ treats the tokenizer’s teacher-forcing reconstruction loss as an intrinsic reward, offering a stable and informative signal directly tied to pixel-space fidelity.
Meanwhile, the variational regularization term plays the role of a constraint for keeping the updated policy close to the base AR model, thereby preserving its learned token distribution. 
Unlike STE, which only updates ground-truth tokens, the RL formulation distributes gradients across all sampled token sequences according to their pixel-space rewards. 
This combination enables broader token-space exploration and rapid adaptation toward pixel-level consistency with limited data and compute.

Experiments on class-to-image and text-to-image generation tasks show that VA-$\pi$ is effective and efficient. 
On ImageNet-1K~\citep{imagenet}, post-training LlamaGen-XXL~\citep{llamagen} with VA-$\pi$ substantially enhances visual fidelity and diversity, reducing the FID from 14.36 to 7.65 and increasing IS from 86.55 to 116.70 without classifier-free guidance.
When applied to text-to-image and unified multimodal generation tasks, VA-$\pi$ improves conditioning accuracy and perceptual quality.
All these improvements are achieved within just training on about $1\%$ of the pretrained dataset, without any external reward model, while using only 13.4\% of the compute cost required by conventional free-running RL methods such as AR-GRPO~\citep{yuan2025argrpo}.

The contributions of this work are as follows:
\begin{itemize}
\item We formulate a variational objective that bridges discrete token modeling and pixel-level reconstruction, aligning AR generators directly with the image distribution.
\item Building on this formulation, we propose {VA}-$\boldsymbol{\pi}$, a post-training framework that leverages RL for optimization. 
By treating the reconstruction as a reward, VA-$\pi$ provides direct pixel-level feedback to the generator policy.
\item VA-$\pi$ is highly compute- and data-efficient, requiring only 25 minutes of post-training on 8×A100 GPUs with $1\%$ of pretraining dataset, without relying on external reward models or expensive free-running sampling.
\item VA-$\pi$ consistently improves visual quality across both class-to-image and text-to-image generation tasks, with minimal data and compute, demonstrating a practical path toward efficient alignment of visual AR models.
\end{itemize}
\section{Related Work}
\label{sec:related-work}

\subsection{Auto-Regressive Visual Generation}
Autoregressive (AR) models~\citep{darn, pixelcnn, pixelrnn, imagetransformer, igpt, llamagen} have emerged as a competitive paradigm for visual generation, rivaling diffusion-based~\citep{maskgit, magvit-v2, maskbit} and masked generative models~\citep{adm, dit, sit, repa}.
Direct pixel-level autoregression is computationally prohibitive, so modern AR frameworks use patch-based discrete tokenizers~\citep{vqvae, vqgan}. The tokenizers compress local image regions into latent tokens; recent works focus on mitigating quantization artifacts in the  tokenizers~\citep{magvit-v2, fsq, unitok, mar} and on extending the tokenization beyond regular 2D grids~\citep{titok, onedpiece, flowmo, gigatok}.
However, these advances only refine individual components and overlook a more fundamental bottleneck: the objective is separated between the AR generator’s predictive likelihood training and the tokenizer’s pixel-level reconstruction goal.

\subsection{Tokenizer-Generator Alignment}
The two-stage training pipeline for standard AR visual generation introduces an inherent inconsistency between the generator and the tokenizer.
Prior work has reduced this mismatch from two directions. Generator-centric approaches modify the AR training objective, for instance, by adding noisy context to the token sequences~\citep{he2025rear} or by randomizing the token ordering~\citep{randar, rar}. 
Tokenizer-centric approaches adjust the tokenizer to better accommodate the generator. One approach is to enhance decoder robustness to the generator’s sampled token distributions~\citep{qiu2025robustok}; another is to embed the AR generator's causal dependencies into the token structure~\citep{alitok, flextok, ddt}.
While these strategies reduce the mismatch, they typically require costly retraining and intuitive regularization.  Our methods are more effective based on principled objectives with superior efficiency.

\subsection{Reinforcement Learning in Visual Generation}

Reinforcement learning (RL) has emerged as an effective finetuning strategy for visual generation, enabling enhanced reasoning~\citep{t2ir1,zhang2025survey}, alignment with human preferences~\citep{ddpo, dpok, coca}, and improving controllability or prompt alignment~\citep{yuan2025argrpo, wang2025simplear, chen2025flash}. 
To the best of our knowledge, our work is the first to apply RL to directly optimize AR models with respect to pixel-space reconstruction quality.
\section{Preliminaries}
\label{sec:pre}

\subsection{Visual Autoregressive Models}

State-of-the-art visual AR generation models~\citep{januspro, rar, var, alitok} feature two components:
(1) a \textit{visual tokenizer} that converts an image into a sequence of discrete codes, and
(2) an \textit{autoregressive model} that models and samples these codes.

\noindent The \textbf{Visual Tokenizer} compresses an image into discrete tokens while preserving reconstruction quality. Discrete tokenizers~\citep{vqgan,llamagen,magvit-v2,maskgit} includes an encoder $\mathcal{E}$, a quantizer $\mathcal{Q}$, and a decoder $\mathcal{D}$. Given an image $\mathbf{I}\in\mathbb{R}^{3 \times H \times W}$, the feature dimension $C$, and the token sequence length $N$:
\begin{equation}
{\mathbf{z}} = \mathcal{E}(\mathbf{I}), \quad
\mathbf{x} = \mathcal{Q}({\mathbf{z}}), \quad
\hat{\mathbf{I}} = \mathcal{D}(\mathbf{x}),
\end{equation}
where ${\mathbf{z}} \in \mathbb{R}^{C\times N}$ is the latent feature and $\mathbf{x}\in\{1,\dots,K\}^{N}$ is the \textit{discrete code indices}. The distribution of discrete codes can then be learned via the AR model. 

The tokenizer is optimized using a combination of reconstruction and quantization objectives.
Specifically, the loss typically consists of:
(1) a pixel-wise reconstruction loss $\mathcal{L}_\text{MSE}$,
(2) a perceptual reconstruction loss $\mathcal{L}_\text{p}$ such as LPIPS~\citep{lpips},
and (3) a vector quantization loss $\mathcal{L}_\text{q}$~\citep{vqvae}.
The overall training objective with coefficients $\lambda_\text{p}$ and $\lambda_\text{q}$ is:
\begin{equation}
\mathcal{L}_\text{tok}
=
\underbrace{
\mathcal{L}_\text{MSE}
+ \lambda_\text{p}\mathcal{L}_\text{p}
}_{\text{Reconstruction loss}}
+
\underbrace{
\lambda_\text{q}\mathcal{L}_\text{q}
}_{\text{Quantization loss}}.
\label{eq:tok_train}
\end{equation}
We provide the detailed formulation of each loss term, including the quantization objective, in the Appendix~\ref {app:supp_tokenizer}.

\noindent The \textbf{Autoregressive Model} $\pi_\theta$, parameterized by $\theta$, factorizes the distribution of a sequence $\mathbf{x}_{1:N}$ into conditional probabilities, where each token $x_i$ depends on all preceding tokens. The model is trained using teacher forcing, where the ground-truth preceding tokens from the observed sequence are provided as context. Formally, the training maximizes the log-likelihood of the observed sequence:
\begin{equation}
\label{eq:ar_obj}
\theta
= \argmax_\theta \sum_{i=1}^{N}
\log \pi_\theta(\mathbf{x}_i \mid \mathbf{x}_{1:i-1}) .
\end{equation}

During inference, tokens are generated sequentially in free-running mode by sampling $\mathbf{x}_i \sim \pi_\theta(\cdot \mid \mathbf{x}_{1:i-1})$. The complete sequence $\mathbf{{x}}$ is then mapped back to the image space via the decoder
$\mathcal{D}$ to obtain the synthesized image $\hat{\mathbf{I}}$.
In the cases of class- and text-to-image generation, the class or text input would serve as an additional conditioning variable, which we omit from the notation above for simplicity.

\subsection{Reinforcement Learning}
An autoregressive generation process can be formulated for a reinforcement learning (RL) problem, where the AR model $\pi_\theta$ serves as the policy that sequentially samples tokens to maximize an expected reward. This formulation enables optimization of $\pi_\theta$ for given rewards via policy-gradient methods such as PPO~\citep{ppo} or GRPO~\citep{grpo}. We adopt GRPO~\citep{grpo} for its stability, as it normalizes rewards across sample groups to reduce variance and leverages KL-regularization to preserve pretrained priors.

\noindent\textbf{Group Relative Policy Optimization (GRPO).}
In GRPO, for each condition $\mathbf{q}$, the policy $\pi_\theta$ generates a group of $G$ samples ${\mathbf{o}_1, \dots, \mathbf{o}_G}$. Each sample is scored by a reward function $\mathcal{R}(\mathbf{o}, \mathbf{q})$, yielding scalar rewards $r_i = \mathcal{R}(\mathbf{o}i, \mathbf{q})$. To enhance training stability, these rewards are normalized within the group to obtain the group-relative advantages:
\begin{equation}
\label{eq:adv}
\hat{A}_i
=
\frac{r_i - \mathrm{mean}(\{r_j\}_{j=1}^{G})}
{\mathrm{std}(\{r_j\}_{j=1}^{G})},
\end{equation}

The GRPO objective has two terms: a clipped policy-ratio objective, weighted by normalized advantages $\hat{A_i}$ from Eq.~\ref{eq:adv}, and a KL-divergence penalty for stability:

\vspace{-5pt}
\begin{equation}
\label{eq:grpo}
\begin{aligned}
\mathcal{J}_{\text{GRPO}}(\theta)
&=
\mathbb{E}_{\mathbf{q},\,\mathbf{o}\sim\pi_{\theta_{\text{old}}}}
\bigg[
\frac{1}{G}\!\sum_{i=1}^{G}
\min\!\big(\rho_i\hat{A}_i,\,
\\[-3pt]
&\quad\operatorname{clip}(\rho_i,1\!-\!\epsilon,1\!+\!\epsilon)\hat{A}_i\big)
-\;
\beta\, D_{\mathrm{KL}}\!\big(\pi_\theta \,\|\, \pi_{\text{ref}}\big)
\bigg],
\end{aligned}
\end{equation}

\noindent
Here, $\rho_i = \tfrac{\pi_\theta(\mathbf{o}_i \mid \mathbf{q})}{\pi_{\theta_{\text{old}}}(\mathbf{o}_i \mid \mathbf{q})}$ 
is the policy ratio, $\hat{A}_i$ denotes the group-normalized advantage defined in Eq.~\ref{eq:adv}, 
and $\beta$ controls the KL regularization strength. 
The first term encourages \emph{policy improvement}, while the second term acts as a \emph{stability constraint}, 
penalizing large deviations from the reference policy $\pi_{\text{ref}}$, 
thereby preventing policy collapse.

In AR generation, the policy $\pi_\theta$ acts as a token-level generator that sequentially predicts visual tokens. GRPO therefore enhances AR generation quality by optimizing the policy to maximize perceptual or semantic rewards.

\begin{table}[!t]
\centering
\small
\setlength{\tabcolsep}{5pt}
\renewcommand{\arraystretch}{1.15}
\caption{
\textbf{Comparison on text–image alignment metrics.} VA-$\pi$ without reward model attains higher scores than AR-GRPO, even on the alignment reward that AR-GRPO itself is optimized for.
}
\vspace{-1pt}
\scalebox{0.92}{
\begin{tabular}{lccc}
\toprule
\textbf{Model} & \textbf{Ext. Rwd} & \textbf{CLIP$\uparrow$} & \textbf{HPS~v2$\uparrow$} \\
\midrule
LlamaGen-XL & -- & 0.245 & 0.153 \\
\quad + AR-GRPO~\citep{yuan2025argrpo} & \checkmark & 0.274 & 0.208 \\
\quad + VA-$\pi$ (Ours) & $\times$ & \cellcolor{lightblue}{0.291} & \cellcolor{lightblue}{0.211} \\
\bottomrule
\end{tabular}
}
\label{tab:alignment_metrics}
\end{table}

\section{VA-\texorpdfstring{$\boldsymbol{\pi}$}{pi}: Variational Policy Alignment}
\label{sec:pipe}

We propose \textbf{VA}-$\boldsymbol{\pi}$, a post-training framework that optimizes AR generators for pixel-space distribution alignment as shown in Fig~\ref{fig:method}.
From the intractable pixel-level likelihood, we derive an ELBO that yields two training signals: a pixel-space reconstruction objective and a token-level regularization that preserves the AR prior (Sec.~\ref{subsec:elbo}). The regularization term reduces to a simple next-token prediction loss (Sec.~\ref{subsec:reg}), while the reconstruction term is non-differentiable and is therefore optimized as a reward through reinforcement learning (Sec.~\ref{subsec:reward}). We then adopt GRPO~\citep{grpo} to integrate these two components into a single, stable training procedure, yielding the full \textbf{VA}-$\boldsymbol{\pi}$ algorithm (Sec.~\ref{subsec:policy}).
All notations follow in Sec.~\ref{sec:pre}.

\subsection{Evidence Lower Bound for Alignment}
\label{subsec:elbo}
Let $p_{\text{data}}(\mathbf{I})$ denote the real image distribution. We define the discrete token sequence $\mathbf{x}$ as a latent variable. Our alignment objective maximizes the pixel-space likelihood by decoding token sequences through the tokenizer during post-training,
rather than relying solely on the token-level likelihood of the AR model as in Eq.~\ref{eq:ar_obj}:
\begin{equation}
\label{eq:pixel_mle}
\begin{aligned}
&\max_\theta \;
\mathbb{E}_{\mathbf{I}\sim p_{\text{data}}}
\big[
\log p(\mathbf{I};\theta,\phi)
\big], 
\\
&p(\mathbf{I};\theta,\phi)
= \sum_{\mathbf{x}}
p_{}(\mathbf{I}, \mathbf{x};\theta,\phi)
= \sum_{\mathbf{x}}
p_\phi(\mathbf{I}\mid \mathbf{x})\,\pi_\theta(\mathbf{x}),
\end{aligned}
\end{equation}
where the AR model $\pi_\theta$ defines the likelihood over discrete token sequence, the decoder $\mathcal{D}$'s parameters $\phi$ defines a pixel-space likelihood $p_\phi(\mathbf{I}\mid \mathbf{x})$.

However, directly evaluating Eq.~\ref{eq:pixel_mle} is intractable
because of the integral term\footnote{It marginalizes over all latent token sequences $\mathbf{x}$, i.e., $\sum_{\mathbf{x}} p_\phi(\mathbf{I}\mid\mathbf{x})\,\pi_\theta(\mathbf{x})$, which is generally intractable due to the large discrete space.},
as noted in the VAE framework~\citep{vae}. 
To obtain a tractable surrogate objective similarly, we introduce a variational posterior $q_{\psi,\theta}(\mathbf{x}\mid\mathbf{I})$ learned by AR models that approximates the posterior $p(\mathbf{x}\mid\mathbf{I})$. 
Analogous to how a VAE learns the posterior by reconstructing
the ground-truth image, we learn a discrete posterior over ground-truth token sequences by
training the AR generator under \emph{teacher forcing}.
Given encoder $\mathcal{E}_{\psi}$, quantizer $\mathcal{Q}$ and the AR model $\pi_\theta$, the posterior $q_{\psi,\theta}(\mathbf{x}\mid\mathbf{I})$ is defined as:

\begin{equation}
    q_{\psi,\theta}(\mathbf{x}\mid\mathbf{I}) = \prod_{i=1}^{N}
\pi_\theta(\mathbf{x}_i \mid \mathbf{x^*}_{1:i-1}), \quad \mathbf{x^*} = \mathcal{Q} \big({\mathcal{E}_\psi}(\mathbf{I})\big)
\end{equation}

Here, each token is predicted using the true prefix $\mathbf{x}^*_{1:i-1}$ rather
than the model’s own outputs.
Thus $q_{\psi,\theta}(\mathbf{x}\mid\mathbf{I})$ concentrates on sequences that decode
faithfully back to $\mathbf{I}$, whereas free-running sampling ${\mathbf{x}}_i \sim \pi_\theta(\cdot \mid {\mathbf{x}}_{1:i-1})$ quickly drifts
off the data manifold due to error accumulation. Teacher forcing therefore
offers a stable, low-variance approximation to the posterior. Based on the defined posterior, the evidence lower bound (ELBO) of $p(\mathbf{I};\theta,\psi,\phi)$ is:
\begin{equation}
\begin{aligned}
\label{eq:elbo}
\log p(\mathbf{I};\theta,\psi,\phi)
\ge
&\underbrace{
\mathbb{E}_{q_{\phi,\theta}(\mathbf{x}\mid\mathbf{I})}
\big[
\log p_\psi(\mathbf{I}\mid \mathbf{x})
\big]
}_{\text{reconstruction term}}
\\[4pt]
&\quad -
\underbrace{
\mathrm{KL}\big(q_{\phi,\theta}(\mathbf{x}\mid \mathbf{I}) \,\|\, \pi_\theta(\mathbf{x})\big)
}_{\text{prior regularization term}}.
\end{aligned}
\end{equation}

\noindent which can be derived by the Jensen Inequality~\citep{vae}. (The derivation is provided in Appendix~\ref{app:proof_elbo}.) Since jointly tuning both the AR model and tokenizer can be unstable, we only update the AR generator $\pi_\theta$ while keeping the tokenizer $\phi,\psi$ frozen in our settings.

Maximizing the ELBO offers a principled objective that aligns the AR generator’s token distribution $\pi_\theta(\mathbf{x})$ with the pixel-space likelihood. Specifically, the \textit{reconstruction term} enforces that, given an image and its encoded tokens, the AR model under teacher forcing should generate token sequences capable of reconstructing the original image, thereby providing pixel-level supervision for optimization. The \textit{prior regularization} term preserves the AR model’s original token-level likelihood modeling, ensuring consistency with its pretrained distribution. Theoretical analysis of our learning framework and comparison with VAE~\citep{vae} and standard AR based on VQVAE~\citep{vqvae} are provided in the Appendix~\ref {app:theory}.

\subsection{Regularization with Next Token Prediction}
\label{subsec:reg}
During free-running inference, an AR model samples each token from
its own history (${\mathbf{x}}_i \sim \pi_\theta(\cdot \mid {\mathbf{x}}_{1:i-1})$)
instead of conditioning on the ground-truth prefix used in teacher forcing
(${\mathbf{x}}_i \sim \pi_\theta(\cdot \mid {\mathbf{x}^*}_{1:i-1})$), causing small deviations
to accumulate, known as \emph{exposure bias}~\citep{scheduled_sampling}.

The key insight of the prior regularization term is that minimizing it can be viewed as directly minimizing the \emph{exposure bias}, since the KL term in Eq.~\ref{eq:elbo} measures the discrepancy between the teacher-forced distribution $q_{\psi,\theta}(\mathbf{x}\mid\mathbf{I})$ and the free-running distribution $\pi_\theta(\mathbf{x})$.
While various approaches have been proposed to mitigate exposure bias, 
we follow reAR~\citep{he2025rear} for efficiency by introducing contextual noise 
and applying the next-token prediction loss: 
\begin{equation}
\label{eq:ce}
\mathcal{L}_\text{prior}(\pi_\theta, \mathbf{x}^*, \tilde{\mathbf{x}}^*)
= -\frac{1}{N}\sum_{t=1}^{N} 
\log \pi_\theta(\mathbf{x}^*_t \mid \tilde{\mathbf{x}}^*_{<t}),
\end{equation}
where $N$ denotes the sequence length, and 
$\tilde{\mathbf{x}}^* \sim K_\xi(\cdot \mid \mathbf{x}^*)$ 
represents a dependent variable of $\mathbf{x}^*$ corrupted by a kernel $K_\xi$ with perturbation rate $\xi$. 
Additional theoretical analysis and details of corruption are provided in Appendix~\ref{app:prior}.

\subsection{Learning with Reconstruction Reward}
\label{subsec:reward}
Although the ELBO provides a tractable optimization objective, the reconstruction term is difficult to optimize end-to-end because of several non-differentiable operations.
Both the quantizer $\mathcal{Q}$ and the discrete teacher-forcing sampling block gradient flow, preventing direct back-propagation of pixel-space losses.
The \emph{Straight-Through Estimator (STE)}~\citep{vqvae} addresses the quantization issue by treating the codebook lookup as an identity mapping in the backward pass, allowing gradients from the decoder to reach the generator logits (see Appendix~\ref{app:ste}).
However, token sampling poses an additional challenge: while STE enables gradients through quantization, it does not account for sampling probabilities over the categorical distribution, leaving the overall objective biased.
This mismatch motivates us to apply reinforcement learning instead.
Empirical evidence supporting this analysis is provided in Sec.~\ref{sec:main_results_c2i}.

To resolve this, we formulate the problem as a policy optimization, where the AR model is optimized to produce the token sequence that maximizes the reconstruction reward, i.e., the negative reconstruction loss. Given a reference image $\mathbf{I}$, ground-truth tokens $\mathbf{x}^*=\mathcal{Q}(\mathcal{E}(\mathbf{I}))$, tokens sampled by teacher-forcing $\mathbf{x}\!\sim\!\pi_\theta(\cdot\mid\mathbf{x^*})$, and the decoded image $\hat{\mathbf{I}}=\mathcal{D}(\hat{\mathbf{x}})$, the reconstruction reward is:
\begin{equation}
R(\mathbf{x},\mathbf{x}^*)
=
-\!\big(
\mathcal{L}_{\text{MSE}}(\hat{\mathbf{I}},\mathbf{I})
+\lambda_{\text{p}}\mathcal{L}_{\text{p}}(\hat{\mathbf{I}},\mathbf{I})
\big)
\label{eq:reward}
\end{equation}

\noindent We can then use the reward as the goal for reinforcement learning. To avoid multiple forward of the AR model, we use the noisy token sequence $\tilde{\mathbf{x}}^* \sim p_\xi(\cdot \mid \mathbf{x}^*)$ as the same used in next-token prediction regularization. Intuitively, maximizing such reward guides $\pi_\theta$ to produce token sequences whose decoded images align with the reference images, maximizing the reconstruction term in Eq.~\ref{eq:elbo}.

\begin{table*}[!t]
\centering
\small
\setlength{\tabcolsep}{5pt}
\renewcommand{\arraystretch}{1.15}
\caption{
\textbf{Quantitative results on class-conditional ImageNet-1k~\citep{imagenet}.} 
We compare both \textbf{LlamaGen-XL} (775M) and \textbf{LlamaGen-XXL} (1.4B) models. 
All models are evaluated both with and without classifier-free guidance (CFG). 
Generated $384\times384$ images are resized to $256\times256$ for evaluation. 
Metrics include Fréchet Inception Distance (FID), Inception Score (IS), 
Precision (Pre.) and Recall (Rec.). 
``Ext.~Rwd'' denotes the use of external reward during reinforcement learning fine-tuning. 
Our proposed \textbf{VA-$\boldsymbol{\pi}$} achieves competitive diversity (FID) and perceptual quality (IS) with substantially lower training cost. 
Best FID and IS results are highlighted in \colorbox{lightblue}{blue}.
}
\vspace{-1pt}
\scalebox{0.9}{
\begin{tabular}{lcccccccccccc}
\toprule
\multirow{2}{*}{\textbf{Model}} & 
\multirow{2}{*}{\textbf{Ext. Rwd}} & 
\multirow{2}{*}{\textbf{Time (min)$\downarrow$}} &
\multicolumn{4}{c}{\textbf{w/o cfg}} &
\multicolumn{4}{c}{\textbf{w/ cfg}} \\
\cmidrule(lr){4-7} \cmidrule(lr){8-11}
 & & & \textbf{FID$\downarrow$} & \textbf{IS$\uparrow$} & \textbf{Pre.$\uparrow$} & \textbf{Rec.$\uparrow$} 
 & \textbf{FID$\downarrow$} & \textbf{IS$\uparrow$} & \textbf{Pre.$\uparrow$} & \textbf{Rec.$\uparrow$} \\
\midrule
\textbf{LlamaGen-XL (775M)}~\citep{llamagen} & -- & -- & 15.55 & 79.16 & 0.62 & 0.69 & \cellcolor{lightblue}{2.79} & 286.88 & 0.84 & 0.54 \\
\quad + AR-GRPO~\citep{yuan2025argrpo} & \checkmark & 149 & -- & -- & -- & -- & 3.63 & 293.07 & 0.86 & 0.48 \\
\quad + VA-$\pi$ (Ours) & $\times$ & 20 & \cellcolor{lightblue}{9.23} & \cellcolor{lightblue}{111.59} & 0.71 & 0.59 & 2.94 & \cellcolor{lightblue}{299.63} & 0.84 & 0.53 \\
\midrule
\textbf{LlamaGen-XXL (1.4B)}~\citep{llamagen} & -- & -- & 14.36 & 86.55 & 0.63 & 0.69 & 2.37 & 252.16 & 0.81 & 0.59 \\
\quad + Post-train Tokenizer & $\times$ & 18 & 14.26 & 86.70 & 0.63 & 0.68 & 2.72 & 246.97 & 0.80 & 0.59 \\
\quad + Post-train Tokenizer (longer) & $\times$ & 207 & 22.99 & 72.49 & 0.56 & 0.68 & 4.31 & 221.57 & 0.75 & 0.58 \\
\quad + STE based Post-train AR~\citep{vqvae} & $\times$ & 381 & 11.46 & 102.21 & 0.68 & 0.61 & 4.17 & 267.34 & 0.83 & 0.51 \\
\quad + VA-$\pi$ (Ours) & $\times$ & 25 & \cellcolor{lightblue}{7.65} & \cellcolor{lightblue}{116.70} & 0.71 & 0.64 & \cellcolor{lightblue}{2.28} & \cellcolor{lightblue}{273.53} & 0.83 & 0.56 \\
\bottomrule
\end{tabular}
}
\label{tab:c2i_main}
\end{table*}
\begin{table*}[!ht]
\centering
\small
\setlength{\tabcolsep}{5pt}
\renewcommand{\arraystretch}{1.15}
\caption{
\textbf{Quantitative results on the GenEval benchmark.} 
The upper block reports performance of \textbf{LlamaGen-XL} (T2I visual generation model), 
and the lower block reports \textbf{Janus Pro-1B} (unified multi-modal model). 
The abbreviation "Ext. Rwd" denotes "External Reward", "Attr. Bind." denotes "Attribute Binding", "obj." denotes "Object". 
VA-$\pi$ improves over both LlamaGen-XL~\citep{llamagen} and AR-GRPO~\citep{yuan2025argrpo}, achieving the highest overall GenEval~\citep{geneval} score.
When applied to the unified multimodal model Janus-Pro 1B~\citep{team2024chameleon}, VA-$\pi$ further enhances fine-grained attributes, demonstrating its generalization across model architectures.
Best results are highlighted in \colorbox{lightblue}{blue}.
}
\vspace{-1pt}
\scalebox{0.9}{
\begin{tabular}{lcccccccc}
\toprule
\textbf{Model} & \textbf{Ext. Rwd} &
\textbf{Position$\uparrow$} & \textbf{Color$\uparrow$} & \textbf{Attr. Bind.$\uparrow$} & \textbf{Counting$\uparrow$} & \textbf{Single Obj.$\uparrow$} & \textbf{Two Obj.$\uparrow$} & \textbf{Overall$\uparrow$} \\
\midrule
\textbf{LlamaGen-XL}~\cite{llamagen} & -- 
& 0.042 & 0.550 & 0.032 & 0.197 & 0.750 & 0.263 & 0.306 \\
\quad + AR-GRPO~\cite{yuan2025argrpo} & \checkmark 
& 0.040 & 0.593 & 0.030 & 0.228 & \cellcolor{lightblue}{0.791} & 0.263 & 0.324 \\
\quad + VA-$\pi$ (Ours) & $\times$ 
& \cellcolor{lightblue}{0.050} & \cellcolor{lightblue}{0.606} & \cellcolor{lightblue}{0.040} & \cellcolor{lightblue}{0.238} & 0.769 & \cellcolor{lightblue}{0.328} & \cellcolor{lightblue}{0.339} \\
\midrule
\textbf{Janus-Pro 1B}~\citep{januspro} & - &
\cellcolor{lightblue}{0.605} & 0.902 & 0.540 & 0.531 & 0.972 & 0.801 & 0.725 \\
\quad + VA-$\pi$ (Ours) & $\times$ & 
0.600 & \cellcolor{lightblue}{0.912} & \cellcolor{lightblue}{0.585} &
\cellcolor{lightblue}{0.540} & \cellcolor{lightblue}{0.988} &
\cellcolor{lightblue}{0.835} & \cellcolor{lightblue}{0.744} \\
\bottomrule
\end{tabular}
}
\label{tab:geneval_results}
\end{table*}

\subsection{VA-\texorpdfstring{$\boldsymbol{\pi}$}{pi} Policy Optimization}
\label{subsec:policy}

Our key insight is that the reconstruction reward (Eq.~\ref{eq:reward}) and regularization with next-token prediction (Eq.~\ref{eq:prior-chain}) are analogous to the goal of policy optimization and KL penalty in reinforcement learning in Eq.~\ref{eq:grpo}. Although the reconstruction reward and regularization term are independent of any specific reinforcement learning framework, we employ GRPO to follow practices~\cite{yuan2025argrpo, grpo}. Specifically, the \textbf{VA}-$\boldsymbol{\pi}$ objective is to maximize:
\begin{equation}
\begin{aligned}
\mathcal{J}_{\text{\textbf{VA}-$\boldsymbol{\pi}$}}(\theta)
&=
\mathbb{E}_{\substack{
I \sim p_\text{data},\;
\mathbf{x^*} = \mathcal{Q}(\mathcal{E}(I)),\\
\tilde{\mathbf{x}}^* \sim p_{\xi}(\cdot \mid \mathbf{x}^*),\;
\{\mathbf{x}_i\}_{i=1}^{G} \sim \pi_{\theta_{\text{old}}}(\cdot \mid \tilde{\mathbf{x}}^*)
}}
\\
&\quad
\bigg[
\frac{1}{G}
\sum_{i=1}^{G}
\min\!\Big(
\rho_i A_i,\;
\operatorname{clip}(\rho_i, 1-\epsilon, 1+\epsilon)\, A_i
\Big)
\\[-2pt]
&\qquad
- \;\beta \,\mathcal{L}_\text{prior}(\pi_\theta, \mathbf{x^*}, \tilde{\mathbf{x}}^*)
\bigg],
\end{aligned}
\label{eq:vapi}
\end{equation}

\noindent
where $p_\text{data}$ is the data distribution of reference images; $G$ is the number of teacher-forced samples per instance;
$\rho_i = \frac{\pi_\theta(\mathbf{x}_i \mid \tilde{\mathbf{x}}^*)}{\pi_{\theta_{\text{old}}}(\mathbf{x}_i \mid \tilde{\mathbf{x}}^*)}$ is the importance ratio;
$A_i$ is the advantage computed from $R(\mathbf{x}_i, \tilde{\mathbf{x}}^*)$ as defined in Eq.~\ref{eq:reward};
$\epsilon$ is the clipping parameter; $\beta$ weights the regularization, and $\xi$ controls the contextual noise level.

\noindent\textbf{Discussion with AR-GRPO.} 
While the motivation of \textbf{VA}-$\boldsymbol{\pi}$ differs from that of standard AR-GRPO~\citep{yuan2025argrpo}, which aims to maximize an externally defined reward, our method offers several additional advantages: 
(1) Unlike the GRPO framework, which requires maintaining a reference model, the prior regularization in \textbf{VA}-$\boldsymbol{\pi}$ introduces no extra storage overhead; 
(2) \textbf{VA}-$\boldsymbol{\pi}$ avoids the cost of additional rollouts used in AR-GRPO, as all terms are derived from teacher-forcing trajectories, thereby improving training efficiency significantly.
(3) Even without an external reward model, \textbf{VA}-$\boldsymbol{\pi}$ achieves better performance than GRPO models specifically trained to maximize these rewards as Tab.~\ref{tab:alignment_metrics}, highlighting the importance of pixel-level alignment.
\begin{figure*}[!t]
    \centering
    \includegraphics[width=.95\linewidth]{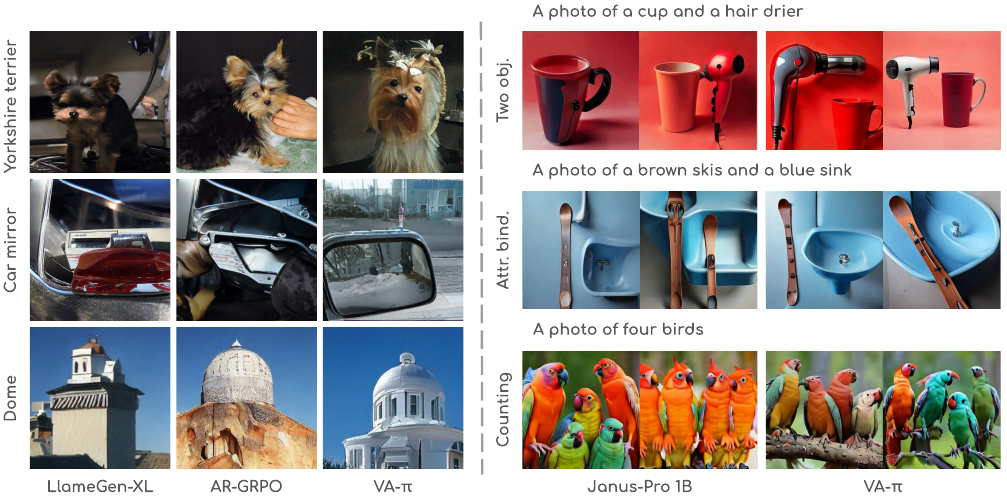}
    \caption{\textbf{Left: Qualitative comparison of C2I generation among LlamaGen-XL~\cite{llamagen} (top), AR-GRPO~\cite{yuan2025argrpo} (middle) and VA-$\boldsymbol{\pi}$ (bottom) on the ImageNet-1k~\citep{imagenet} classes.} Both models use a CFG scale of 2.0. VA-$\pi$ produces clearer object structures (like the car mirror) than LlamaGen-XL (top) and AR-GRPO (middle), demonstrating that pixel-space alignment encourages realistic generations. \textbf{Right: Qualitative comparison of T2I generation between Janus-Pro 1B~\citep{januspro} and VA-$\boldsymbol{\pi}$ on the GenEval Benchmark~\citep{geneval}.} Both models use a CFG scale of 5.0. VA-$\pi$ produces better object combination and counting accuracy, demonstrating stronger capability.}
\vspace{-5pt}
\end{figure*}

\section{Experiments}
\label{sec:experiments}

\subsection{Experimental Setup}
\label{sec:exp_setup}

To validate the effectiveness of VA-$\pi$, we evaluate it on two visual generation tasks: (i) class-conditioned image generation (C2I) and (ii) text-conditioned image generation (T2I). 
For C2I, we adopt LlamaGen~\citep{llamagen} as the base autoregressive (AR) generator. 
For T2I, we assess VA-$\pi$ on both LlamaGen and Janus-Pro~1B~\citep{januspro}, a unified multimodal model (UMM) capable of both understanding and generation. 

In the C2I setting, we employ LlamaGen-XXL (1.4B) and LlamaGen-XL (775M), continuing training on the ImageNet-1k~\citep{imagenet} dataset for 100 steps using 12.8K samples. Each image–text pair produces eight samples for group-wise policy optimization. 
All training is conducted without classifier-free guidance (CFG) to ensure diverse yet stable exploration. 
For the T2I setting, we further fine-tune LlamaGen-XL on the LAION-COCO~\citep{laioncoco} dataset for 200 steps, following the same setup as in C2I except for the noise perturbation ratio, detailed in Section~\ref{sec:ab_study}. 
For the UMM architecture, we fine-tune Janus-Pro~1B on the Flux-Reason dataset~\citep{fluxreason6m} for 100 steps. 
Additional implementation details are provided in Appendix~\ref{app:imp_details}. 

\noindent\textbf{Evaluation metrics.} 
For C2I, we assess image fidelity and diversity using Fréchet Inception Distance (FID) and Inception Score (IS). 
FID quantifies the distributional gap between real and generated images, directly reflecting the alignment objective of VA-$\pi$. 
For T2I, we adopt the GenEval benchmark~\citep{geneval}, which evaluates models on six compositional dimensions: object relations, counting, spatial layout, color, shape, and attribute binding.


\subsection{Main Results on Class-to-Image Generation}
\label{sec:main_results_c2i}

We evaluate VA-$\pi$ on the C2I generation task using ImageNet-1k's validation set (50,000 images). As shown in Tab.~\ref{tab:c2i_main}, VA-$\pi$ is compared with strong baselines built upon the LlamaGen architecture: (i) AR-GRPO~\citep{yuan2025argrpo}: RL finetuned method that aligns the model using multiple external reward models; (ii) Tokenizer-centric approaches: We post-train the tokenizer for 100 steps as VA-$\pi$ and 10,000 steps (longer) on ImageNet-1k~\cite{imagenet}; (iii) Generator-centric approaches: We facilitate gradients flowing back to the generator na\"{i}vely using STE~\cite{vae}.

Post-trained based on LlamaGen-XL (775M), VA-$\pi$ delivers remarkable gains without classifier-free guidance, boosting both FID (15.55 $\!\rightarrow\!$ 9.23) and IS (79.16
 $\!\rightarrow\!$ 111.59). With classifier-free guidance ($scale=2.0$), it achieves the highest IS of 299.63, surpassing AR-GRPO while requiring no external reward model and \textbf{7.5 $\!\times\!$} faster training 
 (20 minutes). For the larger LlamaGen-XXL (1.4B), VA-$\pi$ reduces $\!\sim\!\!50\%\!$ FID (14.35 $\!\rightarrow\!$ 7.65) and increases IS by $30.16$ with only 25 minutes of post-training, outperforming naive post-training strategies such as post-train tokenizer and post-train tokenizer using STE. With classifier-free guidance ($scale=1.75$), it achieves the best FID of 2.28 and IS of 273.53, with 15$\times$ faster training time than STE. VA-$\pi$ uniquely improves both fidelity (FID) and diversity (IS) simultaneously, unlike prior methods that enhanced one at the expense of the other.
Except for the LlamaGen-XL with CFG variant (reasonable since VA-$\pi$ is trained without CFG), it achieves the significant improvement with the lowest training cost.

\subsection{Main Results on Text-to-Image Generation}
\label{sec:main_results_t2i}

We further evaluate VA-$\pi$ on the text-to-image (T2I) generation task using the GenEval~\cite{geneval} benchmarks. As shown in Table~\ref{tab:geneval_results}, we implement our post-train method on both the visual generation model LlamaGen-XL~\citep{llamagen} and the unified multi-modal model Janus-Pro 1B~\citep{januspro}.
VA-$\pi$ achieves consistent improvements, while it does not train on any text-alignment or human preference rewards.

\noindent\textbf{Results on LlamaGen-XL.} On GenEval, our method outperforms AR-GRPO across most sub-tasks, improving the overall score (0.324 $\rightarrow$ 0.339), improved by 0.015. Clear gains are observed on semantically complex prompts like color understanding (+0.013), counting (+0.010), and two-object composition (+0.065). 
Furthermore, we evaluate VA-$\pi$ using CLIP~\citep{clip} on Table~\ref{tab:alignment_metrics} and HPS v2~\citep{hpsv2} with DrawBench~\citep{drawbench} prompts.
VA-$\pi$ outperforms AR-GRPO even without explicit fine-tuning on these evaluation metrics, whereas AR-GRPO requires task-specific adaptation.
This demonstrates the strong generalization ability achieved by our pixel-level alignment strategy.

\noindent\textbf{Results on Janus-Pro 1B.} 
When applied to the unified multi-modal model, VA-$\pi$ further enhances visual compositionality and semantic grounding, 
raising the overall score from 0.725 to 0.744. 
Improvements are particularly prominent in attribute binding (+0.045) and two-object relations (+0.034), 
indicating that VA-$\pi$ generalizes effectively to large multi-modal systems. 
This suggests that the proposed alignment objective provides a scalable mechanism for bridging token-level and perceptual-level consistency in text-conditioned generation.




\begin{figure}[!t]
    \centering
    \begin{minipage}[b]{0.48\linewidth}
        \centering
        \includegraphics[width=\linewidth]{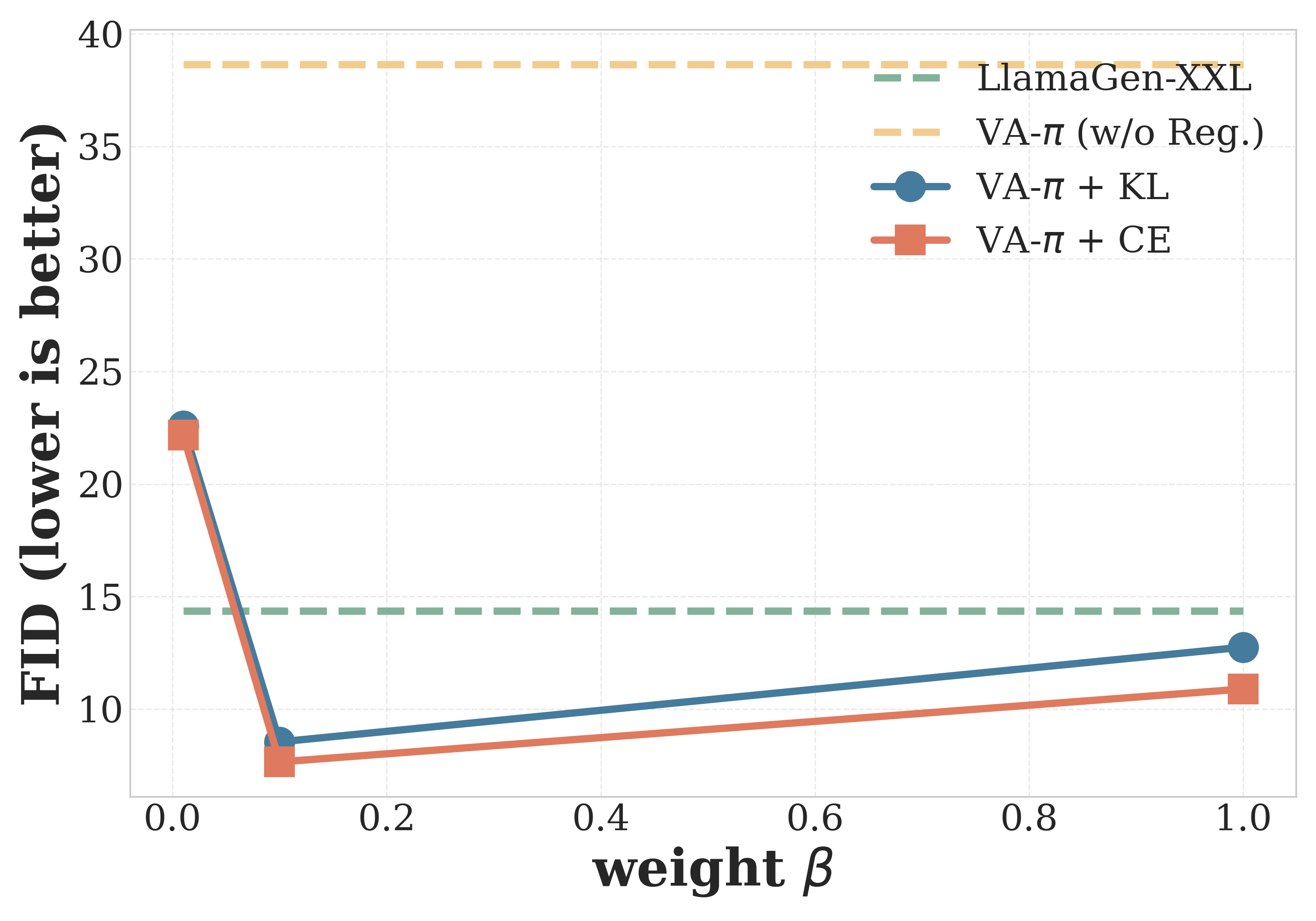}
        \small (a) FID over weight $\beta$
    \end{minipage}
    \hfill
    \begin{minipage}[b]{0.48\linewidth}
        \centering
        \includegraphics[width=\linewidth]{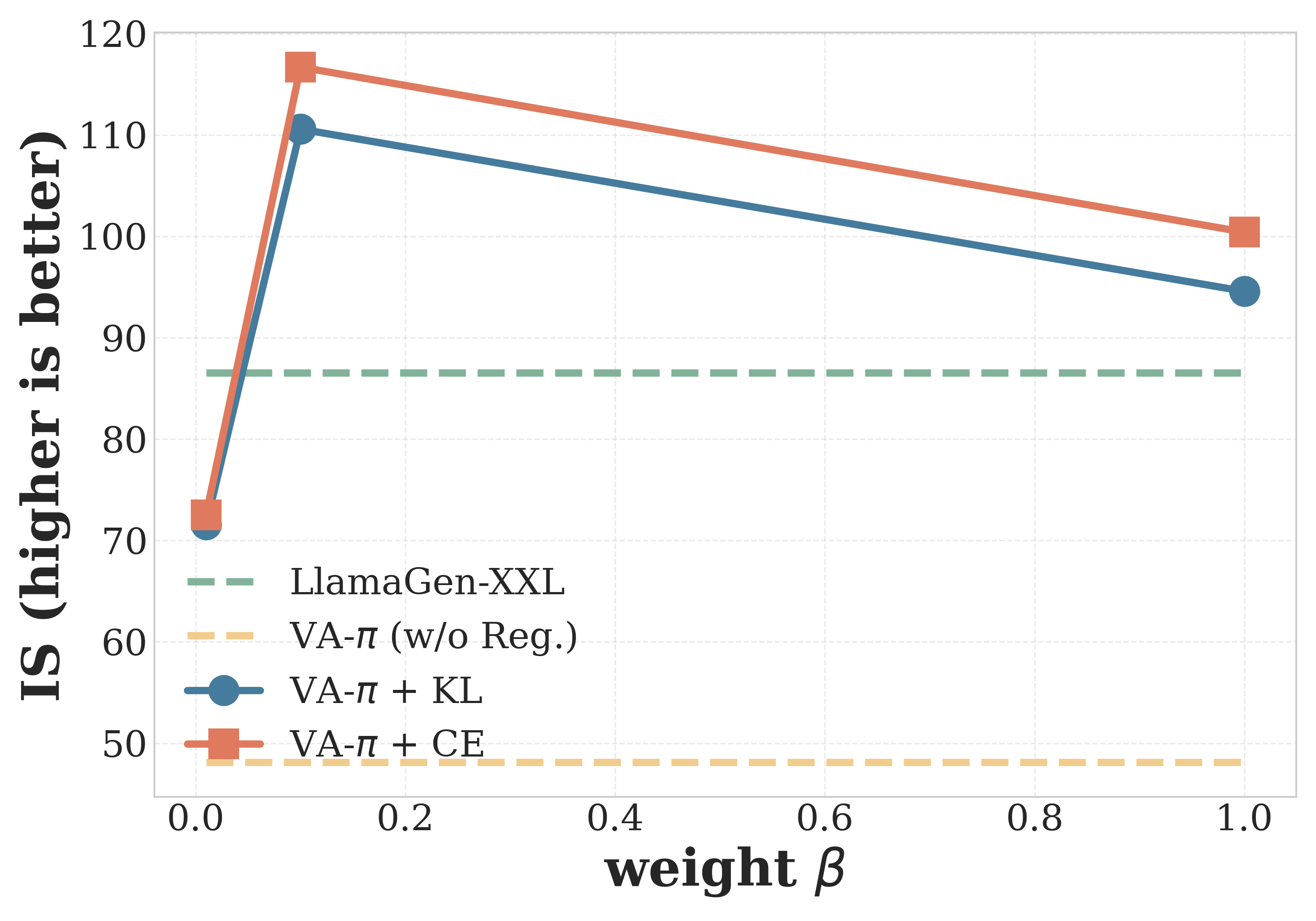}
        \small (b) IS over weight $\beta$
    \end{minipage}
    \caption{\textbf{Ablation on regularization weight (w/o cfg).} CE regularization consistently outperforms KL regularization on FID and IS. Moderate CE regularization ($0.1$) provides the best results.}
    \label{fig:ablation_fid_is}
\end{figure}

\subsection{Ablation Study}
\label{sec:ab_study}

We conduct extensive ablation studies to examine the effectiveness of each component in VA-$\pi$, focusing on reward composition, prior regularization, and Contextual Noise. 

\noindent\textbf{Reward and Loss Composition.}
Table~\ref{tab:ablation_reward} investigates how different reward components (in Sec.~\ref{subsec:reward}) influence training dynamics and generation quality. 
Using only the reconstruction reward ($\mathcal{L}_{p}$ / $\mathcal{L}_{MSE}$) as in Eq.~\ref{eq:reward} fails to provide meaningful alignment 
due to drifting away from the pre-trained AR token distribution without prior regularization constraints as in Eq.~\ref{eq:elbo}.
Incorporating the prior regularization term as an auxiliary objective (i.e., cross-entropy loss) significantly stabilizes optimization by maintaining the token-level likelihood that was already learned in the original AR model. 
The full combination achieves the best balance. This demonstrates that both reconstruction rewards and the prior regularization term are essential for generator–tokenizer consistency.

\noindent\textbf{Prior Regularization Term.}
Fig.~\ref{fig:ablation_fid_is} analyzes the effect of regularization strength (in Sec.~\ref{subsec:reg}) across KL- and CE-based variants.
the regularization strength is defined by $\beta$ as in Eq.~\ref{eq:vapi}
Here, KL refers to Kullback–Leibler divergence regularization, which penalizes deviations from the base policy distribution,
while CE denotes cross-entropy regularization, which enforces consistency with target token probabilities.
Without regularization, optimization diverges rapidly, yielding poor fidelity (FID 38.63). 
Moderate weights ($\beta=0.1$) effectively constrain policy updates, improving both FID and IS. Excessively strong regularization ($\beta=1.0$) oversmooths gradients and suppresses diversity. 
Notably, CE regularization outperforms KL regularization under the same setting. This confirms that lightweight CE regularization ($\beta=0.1$) achieves an optimal trade-off between stability and expressiveness.

\noindent\textbf{Contextual Noise.}
Table~\ref{tab:ablation_noise} examines the impact of stochastic contextual noise to mitigate exposure bias during policy updates (in Sec.~\ref{subsec:reg}). 
We ablate the corruption probability $\xi$ (in Eq.~\ref{eq:ce}) from 0 to 0.95 in the LlamaGen T2I post-train setting.
Injecting a moderate amount of noise ($\xi=0.5$) yields the best overall performance on GenEval (Overall 0.339). In contrast, either no noise ($\xi=0$) or excessive perturbation ($\xi>0.75$) leads to suboptimal results.

\begin{table}[!t]
\centering
\small
\setlength{\tabcolsep}{5pt}
\renewcommand{\arraystretch}{1.15}
\caption{
\textbf{Ablation on reward composition (w/o CFG).} 
We analyze the contribution of each reward component: 
$\mathcal{L}_{\text{MSE}}$ (pixel-level reconstruction), 
$\mathcal{L}_{\text{p}}$ (perceptual similarity via LPIPS~\citep{lpips}), 
and $\mathcal{L}_{\text{prior}}$ (token-level cross-entropy regularization). 
}
\vspace{-1pt}
\begin{tabular}{ccc|cccc}
\toprule
$\mathcal{L}_{\text{MSE}}$ & $\mathcal{L}_{\text{p}}$ & $\mathcal{L}_{\text{prior}}$ & \textbf{FID}$\downarrow$ & \textbf{IS}$\uparrow$ & \textbf{Pre.}$\uparrow$ & \textbf{Rec.}$\uparrow$ \\
\midrule
 &  &  & 14.36 & 86.55 & 0.63 & 0.69 \\
  & \checkmark &  & 38.76 & 49.78 & 0.48 & 0.46 \\
 \checkmark & \checkmark &  & 38.63 & 48.14 & 0.49 & 0.46 \\
  &  & \checkmark & 14.17 & 88.78 & 0.63 & 0.69 \\
 \checkmark & \checkmark & \checkmark & 7.65 & 116.70 & 0.68 & 0.64 \\
\bottomrule
\end{tabular}
\label{tab:ablation_reward}
\end{table}

\begin{table}[!t]
\centering
\small
\setlength{\tabcolsep}{5pt}
\caption{\textbf{Ablation on noise ratio ($\xi$) during training.} 
Moderate noise ratio ($0.5$) achieves the best overall performance on GenEval. 
Abbreviations: PT (Position), CL (Color), AB (Attribute Binding), CT (Counting), SO (Single Object), TO (Two objects).}
\vspace{-1pt}
\scalebox{0.9}{
\begin{tabular}{cccccccc}
\toprule
\textbf{$\boldsymbol{\xi}$} & \textbf{PT} & \textbf{CL} & \textbf{AB} & \textbf{CT} & \textbf{SO} & \textbf{TO} & \textbf{Overall} \\
\midrule
0 & 0.048 & 0.566 & 0.023 & 0.159 & 0.688 & 0.326 & 0.302 \\
0.25 & 0.043 & 0.598 & 0.025 & 0.215 & 0.700 & 0.306 & 0.315 \\
0.5  & 0.050 & 0.606 & 0.040 & 0.238 & 0.769 & 0.328 & 0.339 \\
0.75 & 0.075 & 0.641 & 0.028 & 0.163 & 0.750 & 0.333 & 0.332 \\
0.95 & 0.043 & 0.652 & 0.040 & 0.181 & 0.728 & 0.328 & 0.329 \\
\bottomrule
\end{tabular}
}
\label{tab:ablation_noise}
\end{table}
\section{Conclusion}

We propose \textbf{VA-$\pi$}, a principled RL–based post-training framework that aligns visual AR generators to pixel space through a variational objective grounded in probabilistic modeling.
It substantially improves visual fidelity and compositional alignment on both class-conditional and text-conditional generation tasks, while reducing training cost by 86.6\% compared to conventional RL fine-tuning.
The framework also generalizes effectively to large unified multimodal models such as Janus-Pro~1B.
Overall, VA-$\pi$ provides a lightweight and theoretically grounded path toward bridging token-level modeling and pixel-level generation in a scalable multimodal generation scenario.
\clearpage
\setcounter{page}{1}
\maketitlesupplementary

\appendix

\section{Theoretical Details of VA-\texorpdfstring{$\boldsymbol{\pi}$}{pi}}
\label{app:theory}

In this section, we (1) restate the proof of the ELBO with respect to the random variable and posterior defined in Sec.~\ref{app:proof_elbo}; (2) provide additional insights into its relationship with the VAE~\citep{vae} and autoregressive (AR) models based on VQVAE~\citep{vqvae} in Sec.~\ref{app:vae}; and (3) further justify the formulation of the prior regularization term in Sec.~\ref{app:prior}.

\subsection{Proof of Alignment ELBO}
\label{app:proof_elbo}

While the proof of ELBO is already provided in existing literature such as VAE~\citep{vae}. We restate it with posterior based on discrete token sequence as latent variable.

\noindent\textbf{Setup.}
Let $\mathbf{I}$ denote the observed image and $\mathbf{x}\in\mathcal{X}$ the discrete token sequence generated from autoregressive model.
We assume a generative model with parameters $\theta$ as:
\begin{equation}
p_\theta(\mathbf{I},\mathbf{x}) \;=\; p_\theta(\mathbf{I}\mid\mathbf{x})\,p(\mathbf{x}),
\label{eq:joint_factor}
\end{equation}
where $p(\mathbf{x})$ is a prior over token sequences (e.g., the AR prior) and $p_\theta(\mathbf{I}\mid\mathbf{x})$ is a pixel-space likelihood (e.g., the tokenizer decoder composed with a reconstruction distribution).
Our training objective is the marginal log-likelihood:
\begin{equation}
\log p_\theta(\mathbf{I})
\;=\;
\log \sum_{\mathbf{x}\in\mathcal{X}} p_\theta(\mathbf{I},\mathbf{x})
\quad\text{(integral for continuous $\mathbf{x}$),}
\label{eq:mll_eq13}
\end{equation}
which is generally intractable to evaluate directly because summing/integrating over $\mathbf{x}$ is prohibitive.

\noindent\textbf{Introducing a variational posterior.}
Let $q_\phi(\mathbf{x}\mid\mathbf{I})$ be any distribution supported on $\mathcal{X}$ (in practice, produced by teacher forcing so that it is easy to sample from and to evaluate). Multiply and divide the integrand in~\eqref{eq:mll_eq13} by $q_\phi(\mathbf{x}\mid\mathbf{I})$:
\begin{align}
\log p_\theta(\mathbf{I})
&= \log \sum_{\mathbf{x}} 
    q_\phi(\mathbf{x}\mid\mathbf{I})
    \frac{p_\theta(\mathbf{I},\mathbf{x})}{q_\phi(\mathbf{x}\mid\mathbf{I})}.
\label{eq:rewrite}
\end{align}

\noindent\textbf{Jensen's inequality.}
We now apply Jensen's inequality to the concave function $\log(\cdot)$:
\begin{equation*}
\log \mathbb{E}_{q_\phi}\!\big[f(\mathbf{x})\big]
\;\ge\;
\mathbb{E}_{q_\phi}\!\big[\log f(\mathbf{x})\big]
\qquad (\text{for } f(\mathbf{x})>0).
\end{equation*}
Using $f(\mathbf{x}) \!=\! \frac{p_\theta(\mathbf{I},\mathbf{x})}{q_\phi(\mathbf{x}\mid\mathbf{I})}$ in~\eqref{eq:rewrite} gives:
\begin{align}
\log p_\theta(\mathbf{I})
&\ge 
\sum_{\mathbf{x}} q_\phi(\mathbf{x}\mid\mathbf{I})
\log \frac{p_\theta(\mathbf{I},\mathbf{x})}{q_\phi(\mathbf{x}\mid\mathbf{I})}
\label{eq:jensen_step}
\\
&=
\mathbb{E}_{q_\phi(\mathbf{x}\mid\mathbf{I})}\!\big[\log p_\theta(\mathbf{I},\mathbf{x})\big]
-
\mathbb{E}_{q_\phi(\mathbf{x}\mid\mathbf{I})}\!\big[\log q_\phi(\mathbf{x}\mid\mathbf{I})\big]
\label{eq:expand_logs}
\\
&=
\mathbb{E}_{q_\phi(\mathbf{x}\mid\mathbf{I})}\!\big[\log p_\theta(\mathbf{I}\mid\mathbf{x})\big]
+
\mathbb{E}_{q_\phi(\mathbf{x}\mid\mathbf{I})}\!\big[\log p(\mathbf{x})\big] \\
& \quad -
\mathbb{E}_{q_\phi(\mathbf{x}\mid\mathbf{I})}\!\big[\log q_\phi(\mathbf{x}\mid\mathbf{I})\big]
\label{eq:use_factorization}
\\
&=
\mathbb{E}_{q_\phi(\mathbf{x}\mid\mathbf{I})}\!\big[\log p_\theta(\mathbf{I}\mid\mathbf{x})\big]
-
\mathrm{KL}\!\left(q_\phi(\mathbf{x}\mid\mathbf{I}) \,\|\, p(\mathbf{x})\right).
\label{eq:elbo_eq14}
\end{align}
We define the right-hand side of~\eqref{eq:elbo_eq14} as the \emph{evidence lower bound (ELBO)}:
\begin{equation}
\mathcal{L}(\theta,\phi;\mathbf{I})
\;\triangleq\;
\mathbb{E}_{q_\phi(\mathbf{x}\mid\mathbf{I})}\!\big[\log p_\theta(\mathbf{I}\mid\mathbf{x})\big]
-
\mathrm{KL}\!\left(q_\phi(\mathbf{x}\mid\mathbf{I}) \,\|\, p(\mathbf{x})\right),
\end{equation}
so that $\log p_\theta(\mathbf{I}) \ge \mathcal{L}(\theta,\phi;\mathbf{I})$.

\noindent\textbf{Optimization.} When $q_\phi(\mathbf{x}\mid\mathbf{I})$ perfectly matches the true posterior $p_\theta(\mathbf{x}\mid\mathbf{I})$, the KL term becomes zero, and maximizing the ELBO is equivalent to maximum likelihood estimation (MLE) as proved in existing literature~\citep{vae, vqvae, ddpm}.

\begin{table*}[t]
\centering
\scalebox{0.7}{
\begin{tabular}{lcccc}
\toprule
\textbf{Formulation} & \textbf{$q(\mathbf{x}\mid\mathbf{I})$ (Posterior)} & \textbf{$p(\mathbf{x})$ (Prior)} & \textbf{Trainable Modules} & \textbf{Objective} \\
\midrule
\textbf{VAE} &
Continuous Gaussian Distribution &
Gaussian $\mathcal{N}(0,I)$ &
Encoder, Decoder &
Reconstruction + Generation \\

\textbf{VQVAE} &
Categorical Distribution (Dirac) &
Uniform categorical &
Encoder, Decoder, Codebook &
Reconstruction \\

\textbf{AR on VQVAE} &
Categorical Distribution (Dirac) &
AR model &
AR model &
Generation \\

\textbf{VA-$\boldsymbol{\pi}$ (Ours)} &
Teacher-forced Posterior via AR &
AR model &
AR model &
Reconstruction + Generation \\
\bottomrule
\end{tabular}
}
\caption{
Comparison of probabilistic formulations among VAE, VQVAE, VQVAE + AR, and the proposed VA-$\boldsymbol{\pi}$. 
Our method redefines the posterior through teacher-forced AR modeling, maintaining ELBO validity while enabling post-training alignment between the AR generator and the tokenizer.
}
\label{tab:compare_va_pi}
\end{table*}

\subsection{Comparison with VAE and AR on VQVAE}
\label{app:vae}
Our theoretical framework can be justified from the perspective of VAE and AR based on VQVAE. Notice that variational optimization is independent of the choice of the latent variable, posterior and prior. Specifically, the ELBO of \textbf{VA}-$\boldsymbol{\pi}$ can be seem as a variant of VQVAE's ELBO with redefined prior and posterior as shown in Tab~\ref{tab:compare_va_pi}.
\vspace{1pt}

\noindent\textbf{VAE}~\citep{vae} treats the latent variable is continuous feature, with prior in standard Gaussian distribution and the posterior is parameterized by the encoder. During sampling, the model firstly sample noise from standard Gaussian distribution and forward it to decoder to get the generated image.

\noindent\textbf{VQVAE} used in standard visual AR generation~\citep{vqvae, vqgan} instead treats the latent variable as a discrete token sequence, consistent with our formulation. The key distinction, however, lies in how the posterior and prior are defined. In VQVAE, the posterior is determined solely by the encoder and quantizer: given a reference image, the model produces a deterministic token sequence that can be viewed as sampling from a delta distribution, i.e., $q_\phi(\mathbf{x}\mid\mathbf{I})$ is one-hot. The prior, in contrast, is assumed to follow a uniform categorical distribution. Under these assumptions, the KL divergence term becomes a constant independent of learnable parameters and is thus omitted from the objective. However, since the mismatch exists between the assumed uniform prior and the trained empirical posterior, it's challenging to generate realistic samples by decoding directly from the uniform prior. To bridge this gap, an AR model is subsequently trained to learn a more accurate prior over the token space, replacing the uniform assumption in VQVAE during sampling. This two-stage design inherently creates a gap between the AR model and the original VQVAE tokenizer.

While such gap is difficult to be resolved in the pretrained-stage due to the challenging optimization over non-differentiable discrete latent variable, we consider that it can be solved in the post-training settings. Given pretrained autoregressive model, we can reuse it as the prior and redefine the posterior to introduce sampling.

\noindent\textbf{VA}-$\boldsymbol{\pi}$ also represents the latent variable as a discrete token sequence. We redefine the posterior as follows: given a reference image, the encoder and quantizer are used to obtain a deterministic code, similar to the standard VQVAE formulation. However, unlike VQVAE, we further \emph{teacher-force} this code into the AR model to compute a categorical distribution. This design keeps the ELBO theoretically valid, as the bound holds regardless of the specific choice of posterior, while offering two practical advantages.
(1) Since the AR model is incorporated into the posterior, it receives direct supervision from the reconstruction term defined in pixel space.
(2) It makes the KL regularization more interpretable: the KL divergence between the teacher-forcing distribution and the free-running distribution corresponds to the exposure bias, which can be reduced through next-token prediction under noisy ground-truth contexts, as shown in prior works~\citep{he2025rear, qiu2025robustok}.
Redefining the posterior enables post-training of the AR model to better align with the tokenizer.

\subsection{Details of Prior Regularization}
\label{app:prior}

To mitigate the inconsistency between the teacher-forced distribution (conditioned on a dataset) and the free-running distribution, we consider a regularization objective that directly addresses the \emph{exposure bias} problem. While next-token prediction under noisy context has been shown to address exposure bias effectively in existing works~\citep{he2025rear, diffusionforcing}, we provide theoretical formulation below.

\noindent\textbf{Next Token Prediction Regularization.} We define the regularization objective as maximizing:
\[
-\mathrm{KL}\!\left(q_{\phi,\theta}(\mathbf{x}\mid\mathbf{I}) \,\|\, \pi_\theta(\mathbf{x})\right)
\quad \text{w.r.t. } \theta,
\]
since the AR model used for teacher forcing is parameterized by the same model, given $\mathbf{x}^*\sim\mathcal{Q}(\mathcal{E}_\phi(\mathbf{I}))$, the objective is equivalent to minimizing:
\begin{equation}
\mathcal{L}_{\text{prior}}(\theta)
= 
\mathrm{KL}\!\left(\pi_{\theta}(\mathbf{x}\mid\mathbf{x}^*) \,\|\, \pi_\theta(\mathbf{x})\right)
.
\end{equation}

where the AR model
\(\pi_\theta\) factorizes as:
\begin{equation}
\begin{aligned}
\pi_\theta(\mathbf{x}) &= \prod_{t=1}^{N} \pi_\theta(x_t \mid x_{<t}),
\qquad \\
\pi_\theta(\mathbf{x}\mid \mathbf{x}^*) &= \prod_{t=1}^{N} \pi_\theta(x_t \mid x^*_{<t}).
\end{aligned}
\end{equation}

For any AR laws \(P(\mathbf{x})=\prod_t P(x_t\mid x_{<t})\) and
\(Q(\mathbf{x})=\prod_t Q(x_t\mid x_{<t})\), the chain rule for KL gives:
\begin{equation}
\label{eq:kl-chain}
\begin{aligned}
\mathrm{KL}(P\|Q)
= \sum_{t=1}^N \mathbb{E}_{x_{<t}\sim P}\!
\big[
\mathrm{KL}\big(P(\cdot\!\mid\! x_{<t})\,\|\,Q(\cdot\!\mid\! x_{<t})\big)
\big].
\end{aligned}
\end{equation}
We also use the cross-entropy decomposition
\(\mathrm{KL}(p\|q)=\mathcal{H}(p,q)-\mathcal{H}(p)\),
with \(\mathcal{H}(p,q):=-\mathbb{E}_{y\sim p}[\log q(y)]\).

Applying \eqref{eq:kl-chain} to
\(\pi_\theta(\mathbf{x}\mid \mathbf{x}^*)\) vs.~\(\pi_\theta(\mathbf{x})\),
\begin{equation}
\label{eq:prior-chain}
\begin{aligned}
\mathcal{L}_{\text{prior}}(\theta)
&= \sum_{t=1}^{N}
\mathbb{E}_{\mathbf{x}\sim \pi_\theta(\cdot\mid \mathbf{x}^*)}\!
\Big[
\mathrm{KL}\big(\pi_\theta(\cdot\mid x^*_{<t}) \,\|\, \pi_\theta(\cdot\mid x_{<t})\big)
\Big]
\\
&= \sum_{t=1}^{N}
\mathbb{E}_{\mathbf{x}\sim \pi_\theta(\cdot\mid \mathbf{x}^*)}\!
\Big[
\mathcal{H}\big(\pi_\theta(\cdot\mid x^*_{<t}), \pi_\theta(\cdot\mid x_{<t})\big)
\\
& \qquad - \mathcal{H}\big(\pi_\theta(\cdot\mid x^*_{<t})\big)
\Big].
\end{aligned}
\end{equation}

We make two assumptions:
\begin{itemize}[leftmargin=1.2em,itemsep=0pt,topsep=2pt]
\item[\textbf{A1}] (\emph{Teacher-forced calibration})
\(
\pi_\theta(\cdot\mid x^*_{<t}) = p^*(\cdot\mid x^*_{<t})
\) for all \(t,x^*_{<t}\), given $\pi_\theta$ is exactly pretrained on this.
\item[\textbf{A2}] (\emph{Prefix-matching corruption})
There exists a Markov kernel
\(K_\xi(\tilde{x}_{<t}\mid x^*_{<t})\) such that, when
\(\mathbf{x}\sim \pi_\theta(\cdot\mid \mathbf{x}^*)\),
\begin{equation}
\label{eq:prefix-law}
p(x_{<t}\mid \mathbf{x}^*) \;=\; K_t(\cdot\mid x^*_{<t})
\quad \text{for all } t .
\end{equation}
\end{itemize}

Using \eqref{eq:prefix-law} to replace the expectation over
\(x_{<t}\) in \eqref{eq:prior-chain} by an expectation over
\(\tilde{x}_{<t}\sim K_t(\cdot\mid x^*_{<t})\),
and applying A1,
\begin{equation}
\label{eq:ntp-noisy}
\begin{aligned}
\mathcal{L}_{\text{prior}}(\theta)
&=
\sum_{t=1}^{N}
\mathbb{E}_{\tilde{x}_{<t}\sim K_t(\cdot\mid x^*_{<t})}
\mathbb{E}_{y\sim p^*(\cdot\mid x^*_{<t})}
\big[ -\log \pi_\theta(y \mid \tilde{x}_{<t}) \big]
\\[-0.15em]
&\qquad - \sum_{t=1}^N \mathcal{H}\big(p^*(\cdot\mid x^*_{<t})\big).
\end{aligned}
\end{equation}
Let \(C:=\sum_{t}\mathcal{H}(p^*(\cdot\mid x^*_{<t}))\), which is independent of \(\theta\).
Then:
\begin{equation}
\begin{aligned}
\;
\mathcal{L}_{\text{prior}}(\theta)
&\;=\;
\underbrace{\sum_{t=1}^{N}
\mathbb{E}_{\tilde{x}_{<t}\sim K_t}
\mathbb{E}_{y\sim p^*(\cdot\mid x^*_{<t})}
\big[ -\log \pi_\theta(y \mid \tilde{x}_{<t}) \big]}_{\displaystyle
\mathcal{L}_{\text{NTP-noisy}}(\theta)} \\
&\qquad \;-\; C .
\end{aligned}
\end{equation}
Hence, minimizing the prior KL is equivalent up to a constant to minimizing the
next-token prediction (NTP) loss under perturbed prefixes \(\tilde{x}_{<t}\). While the assumption is not guaranteed in the experimental settings, we found it works empirically well by choosing proper corruption kernel.

\noindent\textbf{Corruption Kernel Details}. Following previous work~\citep{he2025rear}, we use uniform noise to implement the corruption kernel. Given a discrete sequence $\mathbf{x}^*=(x_1^*,\dots,x_N^*)$, we define a corruption kernel
$K_\xi(\hat{\mathbf{x}}\mid\mathbf{x}^*)$ that introduces random perturbations with rate $\xi\in[0,1]$.
For each position $i\in\{1,\dots,N\}$, we independently draw:
\begin{equation}
\label{eq:kernel-def}
\hat{x}_i =
\begin{cases}
x_i^*, & \text{with probability } 1-\xi,\\[3pt]
u_i, & \text{with probability } \xi,
\end{cases}
\end{equation}
where $u_i$ is sampled uniformly from the token vocabulary
$\mathcal{V}=\{1,\dots,K\}$ excluding the ground-truth token,
i.e.\ $u_i\!\sim\!\mathrm{Unif}(\mathcal{V}\!\setminus\!\{x_i^*\})$.
Equivalently, the conditional probability mass function is:
\begin{equation}
\label{eq:kernel-mass}
K_\xi(\hat{x}_i\mid x_i^*)=
(1-\xi)\,\delta(\hat{x}_i\!=\!x_i^*)+\xi\,\frac{\mathbf{1}[\hat{x}_i\!\neq\!x_i^*]}{K-1},
\end{equation}
and the full sequence corruption factorizes as
$K_\xi(\hat{\mathbf{x}}\mid\mathbf{x}^*)=\prod_{i=1}^{N}K_\xi(\hat{x}_i\mid x_i^*)$.

\section{Tokenizer Training Objectives}
\label{app:supp_tokenizer}

The visual tokenizer is trained to map continuous image features into a compact set of discrete embeddings while maintaining perceptual reconstruction quality.
As discussed in Sec.~\ref{sec:pre}, the total objective in Eq.~\ref{eq:tok_train} consists of reconstruction and quantization terms:

\begin{equation}
\mathcal{L}_\text{tok}
=
\mathcal{L}_\text{MSE}
+ \lambda_\text{p}\mathcal{L}_\text{p}
+ \lambda_\text{q}\mathcal{L}_\text{q}.
\end{equation}

\noindent\textbf{Pixel-wise reconstruction loss.}
We employ the mean squared error (MSE) between the original image $\mathbf{I}$ and its reconstruction $\hat{\mathbf{I}}$:
\begin{equation}
\mathcal{L}_\text{MSE} = \|\mathbf{I} - \hat{\mathbf{I}}\|_2^2.
\end{equation}

\noindent\textbf{Perceptual reconstruction loss.}
Following~\citep{lpips}, a perceptual loss $\mathcal{L}_\text{p}$ compares high-level feature activations of $\mathbf{I}$ and $\hat{\mathbf{I}}$ extracted by a pretrained VGG network:
\begin{equation}
\mathcal{L}_\text{p} = \sum_l \|\phi_l(\mathbf{I}) - \phi_l(\hat{\mathbf{I}})\|_2^2,
\end{equation}
where $\phi_l(\cdot)$ denotes features from the $l$-th layer.

\noindent\textbf{Vector quantization loss.}
The quantization loss~\citep{vqvae} encourages the encoder output $\mathbf{z}$ to commit to the closest embedding vector in the codebook $\mathcal{E}=\{e_k\}_{k=1}^K$:
\begin{equation}
\mathcal{L}_\text{q}
=
\|\text{sg}[\mathbf{z}] - e\|_2^2
+ \beta \|\mathbf{z} - \text{sg}[e]\|_2^2,
\end{equation}
where $\text{sg}[\cdot]$ denotes the stop-gradient operator and $\beta$ controls the commitment strength.
The first term updates the codebook entries to match encoder outputs, while the second prevents codebook collapse by penalizing large deviations of $\mathbf{z}$ from its assigned embedding.

Together, these terms ensure high-fidelity reconstruction, perceptual realism, and stable codebook learning. 
The resulting tokenizer provides discrete tokens that effectively balance compression and visual quality for subsequent autoregressive modeling.

\section{Implementation Details}
\label{app:imp_details}

We first present the detailed formulation of baselines for mitigating discrepancy between decoding the AR-generated token sequences and the ground-truth image distribution by fine-tuning the AR model by STE algorithm~\ref{app:ste} or fine-tuning the tokenizer decoder~\ref{app:tok_post}. Then, we present the implementation details of our VA=$\pi$ on three experimental setting~ref{}: LlamaGen for C2I and T2I, Janus-Pro for T2I. 

\subsection{Straight-Throught Estimator Variant}
\label{app:ste}

Autoregressive (AR) image generators operate over discrete token sequences produced by a visual tokenizer.
Because the token selection process (\emph{e.g.}, $\arg\max$ over logits) is non-differentiable, directly optimizing pixel-level objectives is intractable.
To enable gradient-based training, we employ the \textit{Straight-Through Estimator (STE)}~\citep{vqvae}, which provides a differentiable surrogate for the discrete sampling process of the AR model.

\noindent\textbf{Forward pass.} Given an image $\mathbf{I}$, the tokenizer encodes it into discrete token indices $\mathbf{x}^* = \mathcal{Q}(\mathcal{E}(\mathbf{I}))$.
During training, the AR generator $\pi_\theta$ predicts the next-token logits $\mathbf{l}_t \in \mathbb{R}^V$ conditioned on the teacher-forced prefix $\mathbf{x}^*_{<t}$:
\[
\mathbf{l}_t = \pi_\theta(\mathbf{x}^*_{<t}).
\]
A temperature-scaled softmax produces a differentiable relaxation of the categorical distribution:
\[
\mathbf{y}_{\text{soft}} = \mathrm{softmax}(\mathbf{l}_t / \tau),
\]
and a discrete token is sampled by a hard one-hot projection by $\arg\max$ as:
\[
\mathbf{y}_{\text{hard}} = \mathrm{onehot}(\arg\max(\mathbf{l}_t)).
\]
The final surrogate variable used for decoding is as:
\[
\mathbf{y}_{\text{st}} = (\mathbf{y}_{\text{hard}} - \mathbf{y}_{\text{soft}}).\mathrm{detach()} + \mathbf{y}_{\text{soft}},
\]
so that the forward path uses $\mathbf{y}_{\text{hard}}$ (discrete tokens), while the backward path reuses the gradient of $\mathbf{y}_{\text{soft}}$.

\noindent\textbf{Backward pass.} Let $\mathbf{e}_k$ denote the codebook embedding corresponding to token index $k$.
In the forward pass, $\mathbf{y}_{\text{hard}}$ selects the embedding $\mathbf{e}_{k^*}$ from the codebook to form the quantized representation $\mathbf{z}_q$.
During back-propagation, the STE treats this quantization step as an identity mapping:
\[
\frac{\partial \mathbf{z}_q}{\partial \mathbf{l}_t} \approx \frac{\partial \mathbf{y}_{\text{soft}}}{\partial \mathbf{l}_t},
\]
which effectively copies the gradient from the decoder output $\mathcal{D}(\mathbf{z}_q)$ to the generator logits $\mathbf{l}_t$.
This allows the AR generator to receive pixel-space reconstruction gradients through the frozen tokenizer without requiring a differentiable relaxation of the discrete sampling.

The STE thus provides a simple yet effective mechanism for propagating pixel-level feedback to the AR generator.
In our implementation, it is applied consistently during the optimization of both reconstruction and regularization objectives.
While the forward path remains discrete for accurate decoding, the backward path transmits continuous gradients to stabilize learning.
This approximation has been widely adopted in discrete generative models~\citep{vqvae} and is crucial for enabling end-to-end optimization. However, STE-based fine-tuning requires more data and training time cost compred to our proposed VA-$\pi$ for the model only update under the ground-truth path.

\subsection{Tokenizer Post-training}
\label{app:tok_post}

A tokenizer trained only on ground-truth image reconstructions may not fully account for the distribution shift induced by AR-generated token sequences. From the tokenizer’s perspective, enhancing robustness to such off-manifold or suboptimal token sequences provides an additional means to reduce the mismatch between AR-generated token distributions and the underlying image distribution.

\noindent\textbf{Formulation.}
Recall that the tokenizer consists of an encoder $\mathcal{E}$, a quantizer $\mathcal{Q}$, and a decoder $\mathcal{D}_{\phi}$.  
Given an image $\mathbf{I} \in \mathbb{R}^{3 \times H \times W}$, the tokenizer produces its latent representation and discrete token sequence:
\begin{equation}
\mathbf{z} = \mathcal{E}(\mathbf{I}), 
\qquad
\mathbf{x}^{*} = \mathcal{Q}(\mathbf{z}),
\end{equation}
where $\mathbf{z} \in \mathbb{R}^{C \times N}$ and $\mathbf{x}^{*} \in \{1,\dots,K\}^{N}$.
The sequence $\mathbf{x}^{*}$ serves as the teacher-forcing target for the AR model, whose parameters are trained by maximizing:
\begin{equation}
\theta
= \argmax_{\theta}
\sum_{i=1}^{N} \log \pi_{\theta}(x^{*}_i \mid \mathbf{x}^{*}_{1:i-1}).
\end{equation}

During tokenizer post-training, we keep the AR model $\pi_{\theta}$, the quantizer $\mathcal{Q}$, and its codebook fixed.  
Given a teacher-forcing sequence $\mathbf{x}^{*}$, the AR model generates its own token predictions autoregressively:
\begin{equation}
x_i \sim \pi_{\theta}(\cdot \mid \mathbf{x}^*_{1:i-1}),
\qquad
\mathbf{x} = (x_1,\dots,x_N),
\end{equation}
which reflect the model’s free-running token distribution conditioned on $\mathbf{x}^{*}$. The AR-generated sequences $\mathbf{x}$ are decoded by the tokenizer decoder:
\begin{equation}
\hat{\mathbf{I}} = \mathcal{D}_{\phi}(\mathbf{x}),
\end{equation}
and the decoder parameters $\phi$ are updated to improve the reconstruction fidelity of AR-generated tokens.  
Because the quantizer $\mathcal{Q}$ and the AR model $\pi_\theta$ remain frozen, the teacher-forcing token distribution $\mathbf{x}^{*}$ is preserved, ensuring compatibility with the pretrained AR model while enhancing robustness to its sampled token sequences.

\noindent\textbf{Objective.} Unlike the full tokenizer objective in Eq.~\ref{eq:tok_train}, the post-training objective excludes the quantization loss and focuses solely on reconstruction fidelity:
\begin{equation}
\mathcal{L}_{\text{PT}}
=
\|\hat{\mathbf{I}} - \mathbf{I}\|_2^2
+
\lambda_{\mathrm{p}}\,
\mathcal{L}_\text{LPIPS}(\hat{\mathbf{I}}, \mathbf{I}).
\end{equation}
This updates only the reconstruction pathway without altering the discrete token space.

The improved tokenizer thus serves as a drop-in replacement that yields higher-fidelity decoding for both ground-truth and AR-generated token sequences without inducing any distributional mismatch.

\subsection{Implementaton Details of VA-$\boldsymbol{\pi}$}
\label{app:imp_vapi}

We summarize the hyperparameters used for RL fine-tuning across all models and tasks in Table~\ref{tab:hyper_c2i_t2i}. 
For both C2I and T2I experiments with LlamaGen, we adopt the same optimizer configuration—AdamW with a learning rate of $1\times10^{-6}$, weight decay of $1\times10^{-4}$, \texttt{bf16} mixed precision, and a global batch size of 128. 
For Janus-Pro 1B, we use a smaller learning rate ($1\times10^{-7}$) due to its increased sensitivity during RL updates. 
All models are trained with group size $G{=}8$, advantage clipping at $5.0$, a classifier-free guidance scale of $1.0$, and a maximum gradient norm of $1.0$.

We fine-tune LlamaGen-XXL for 100 steps on C2I and LlamaGen-XL for 200 steps on T2I, while Janus-Pro 1B is trained for 100 steps given its heavier multimodal architecture. 
Image resolution is set to $384\times384$ for all LlamaGen models and $256\times256$ for Janus-Pro 1B.

\begin{figure*}
    \centering
    \begin{minipage}[b]{0.32\linewidth}
        \centering
        \includegraphics[width=\linewidth]{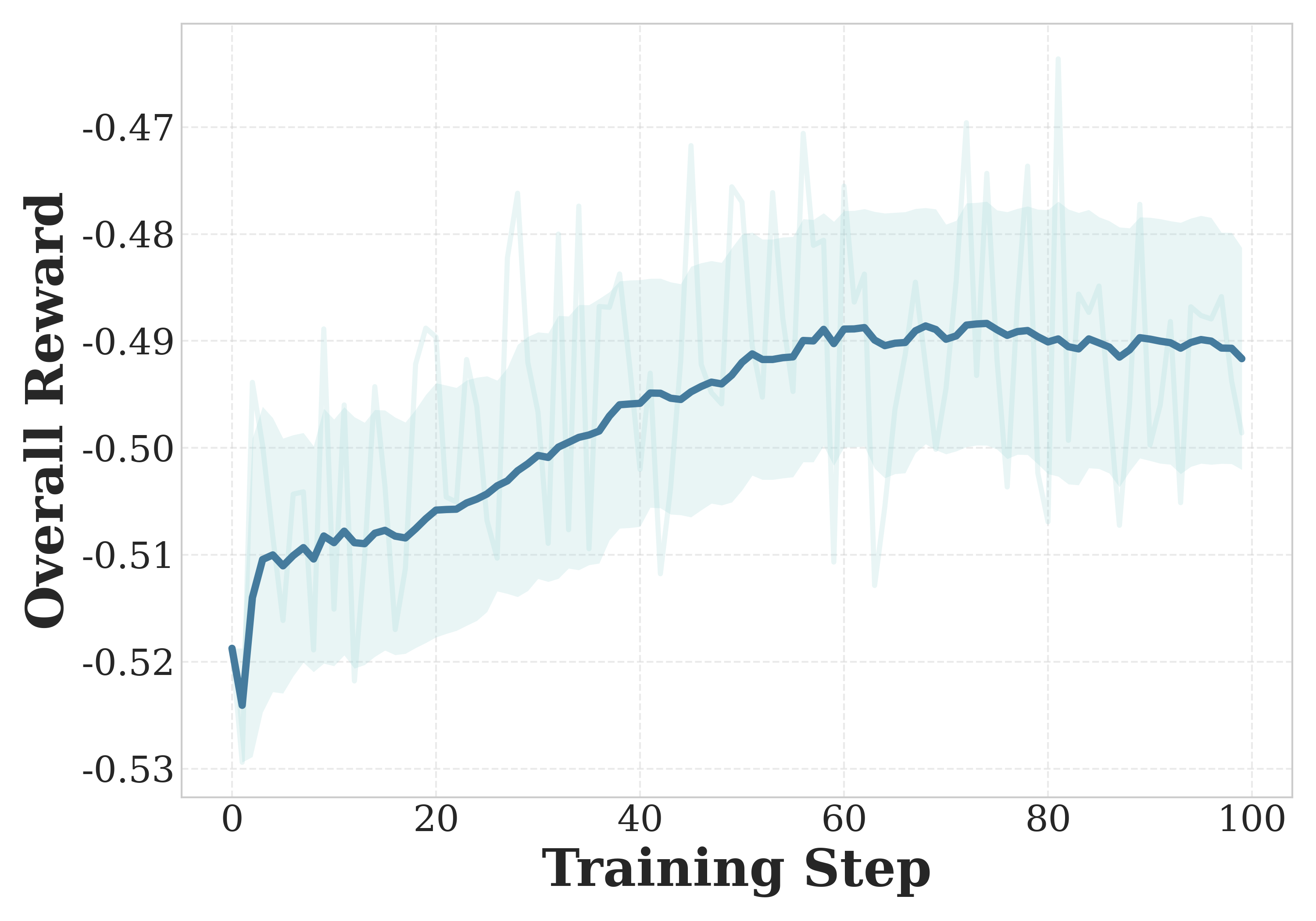}
        \caption*{(a)}
    \end{minipage}
    \hfill
    \begin{minipage}[b]{0.32\linewidth}
        \centering
        \includegraphics[width=\linewidth]{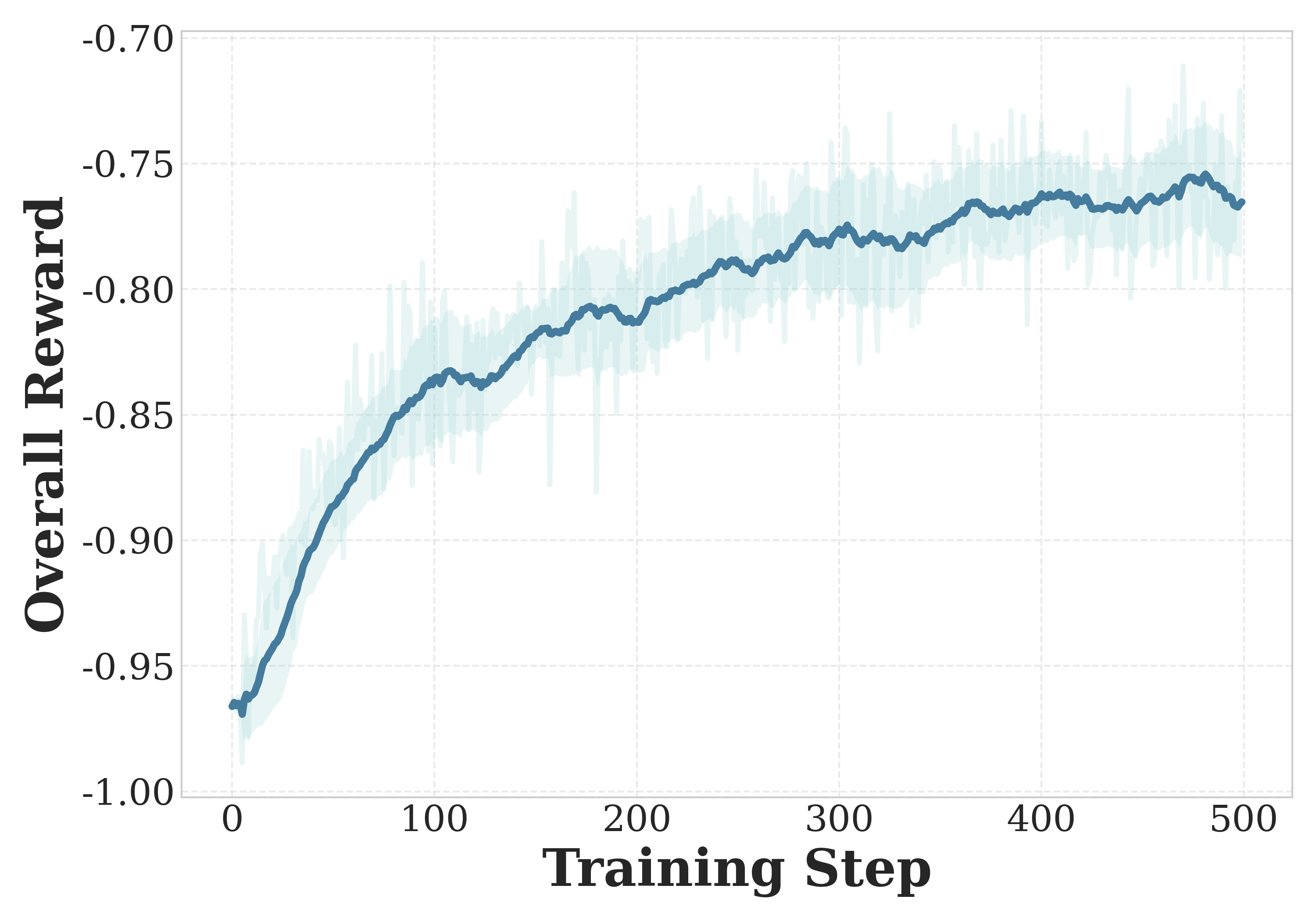}
        \caption*{(b)}
    \end{minipage}
    \hfill
    \begin{minipage}[b]{0.32\linewidth}
        \centering
        \includegraphics[width=\linewidth]{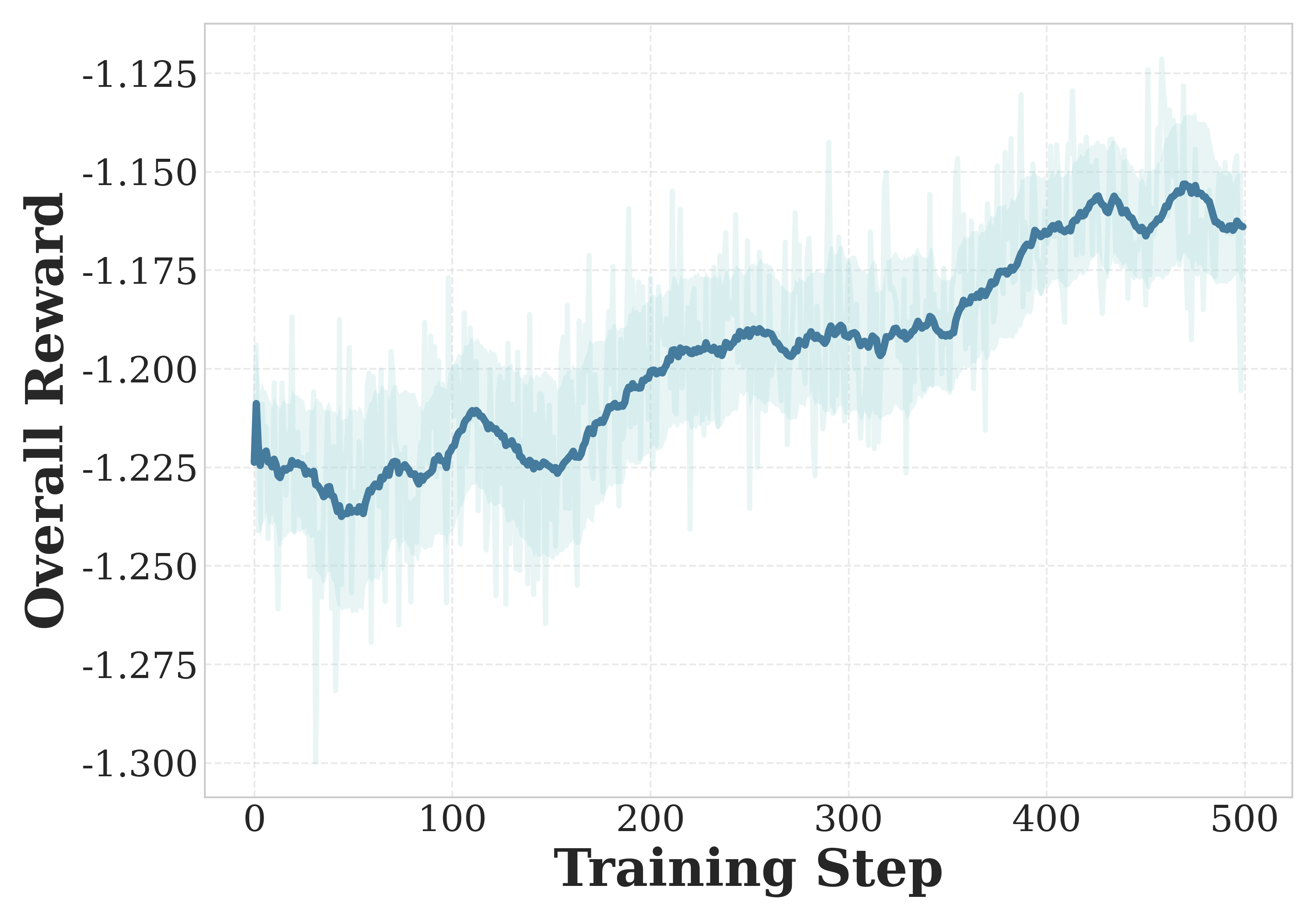}
        \caption*{(c)}
    \end{minipage}
    \caption{\textbf{Learning curves of our reinforcement-learning framework VA-$\boldsymbol{\pi}$ across three model scenarios.} (a) C2I (LlamaGen-XXL, 100 steps), (b) T2I (LlamaGen-XL, 500 steps), and (c) T2I (Janus-Pro 1B, 500 steps).}
    \label{fig:learning_curves_all}
\end{figure*}

\begin{table*}
\centering
\small
\setlength{\tabcolsep}{5pt}
\caption{\textbf{Hyperparameters used for RL fine-tuning across three model settings.} LlamaGen-XXL (C2I) and LlamaGen-XL (T2I) use consistent optimization settings, while Janus-Pro 1B adopts a smaller learning rate and lower resolution due to its multimodal architecture and training stability considerations.}
\label{tab:hyper_c2i_t2i}
\vspace{2mm}
\begin{tabular}{lccc}
\toprule
\textbf{Name} & \textbf{LlamaGen for C2I} & \textbf{LlamaGen for T2I} & \textbf{Janus-Pro 1B for T2I} \\
\midrule
\textbf{Learning Rate} & 1e-6 & 1e-6 & 1e-7 \\
\textbf{Weight Decay} & 1e-4 & 1e-4 & 1e-4 \\
\textbf{Mixed Precision} & bf16 & bf16 & bf16 \\
\textbf{Beta $\boldsymbol{\beta}$} & 0.1 & 0.1 & 0.1 \\
\textbf{Group Size $\boldsymbol{G}$} & 8 & 8 & 8 \\
\textbf{Max Advantage Clip} & 5.0 & 5.0 & 5.0 \\
\textbf{Classifier-Free Guidance Scale} & 1.0 & 1.0 & 1.0 \\
\textbf{Max Gradient Norm} & 1.0 & 1.0 & 1.0 \\
\textbf{Total Batchsize per Step} & 128 & 128 & 128 \\
\textbf{Training Steps} & 100 & 200 & 100 \\
\textbf{Image Resolution $\boldsymbol{h \times w}$} & $384 \times 384$ & $256 \times 256$ & $384 \times 384$ \\
\textbf{Contextual Noise $\boldsymbol{\xi}$} & 0 & 0.5 & 0.95 \\
\bottomrule
\end{tabular}
\end{table*}

\begin{table*}
\centering
\small
\caption{\textbf{Representative prompts used for qualitative comparison on GenEval tasks.}}
\label{tab:geneval-prompts}
\begin{tabular}{l ll}
\toprule
\textbf{Task} & \multicolumn{2}{c}{\textbf{Prompt}} \\
\midrule

\multirow{2}{*}{\textbf{Attribute Binding}} 
 & ``a photo of a brown laptop and a white bicycle'' 
 & ``a photo of a black cow and an orange donut'' \\
 & ``a photo of a blue baseball glove and a black pizza'' 
 & ``a photo of a purple chair and a brown surfboard'' \\
\midrule

\multirow{2}{*}{\textbf{Counting}} 
 & ``a photo of two cups'' 
 & ``a photo of three cell phones'' \\
 & ``a photo of three wine glasses'' 
 & ``a photo of four birds'' \\
\midrule

\multirow{2}{*}{\textbf{Position}} 
 & ``a photo of a sports ball right of a handbag'' 
 & ``a photo of a handbag above a sheep'' \\
 & ``a photo of a skateboard right of a fire hydrant'' 
 & ``a photo of a computer keyboard right of skis'' \\
\midrule

\multirow{2}{*}{\textbf{Two-Object Combination}} 
 & ``a photo of a frisbee and a bottle'' 
 & ``a photo of a book and a tv'' \\
 & ``a photo of a skateboard and a baseball glove'' 
 & ``a photo of a chair and a tv'' \\
\bottomrule

\end{tabular}
\end{table*}

\section{Additional Qualitative Results}
\label{sec:appendix_qualitative}

\subsection{Visualization of Learning Curves}

We visualize the learning curves during reinforcement learning fine-tuning, showing how the reward evolves as training progresses. As illustrated in Fig.~\ref{fig:learning_curves_all}, the reward steadily increases across all settings, indicating that the policy consistently improves under our VA-$\pi$ framework. These curves confirm that our reinforcement learning framework enables robust policy improvement across generators of varying scales and modalities.

\subsection{Class-to-Image Generation}

\begin{figure}
    \centering
    \includegraphics[width=\linewidth]{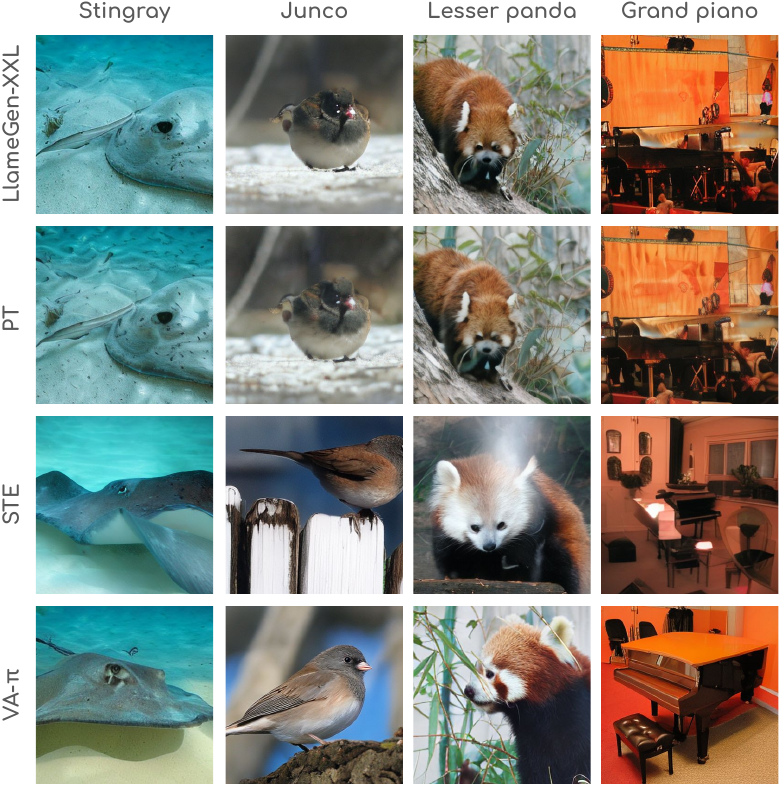}
    \caption{\textbf{Qualitative comparison of C2I generation among LlamaGen-XXL~\cite{llamagen}, post-train tokenizer (PT), STE based post-train AR and VA-$\boldsymbol{\pi}$ on the ImageNet-1k~\citep{imagenet} classes.} Both models use a CFG scale of 1.0. VA-$\pi$ shows better semantic alignment and image quality, demonstrating that pixel-space alignment encourages realistic generations.}
    \label{fig:qual-ste-pt}
\end{figure}

\begin{figure}
    \centering
    \includegraphics[width=\linewidth]{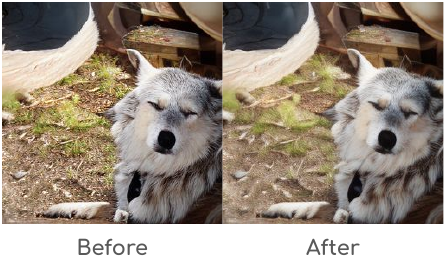}
    \caption{\textbf{Observation on Post-Train Tokenizer.} Post-training the tokenizer decoder produces smoother and less detailed textures, despite preserving global structure. This over-smoothing effect explains why extended decoder fine-tuning leads to worse FID (14.36 $\rightarrow$ 22.99) and IS (86.55 $\rightarrow$ 72.49), highlighting the inherent limitation of decoder-only fine-tuning.}
    \label{fig:qual-pt}
\end{figure}

We present additional qualitative comparisons on the class-to-image (C2I) generation task, showcasing examples from various ImageNet~\citep{imagenet} classes. Each comparison visualizes generations from the baseline LlamaGen-XXL~\citep{llamagen} and our VA-$\pi$, which applies reinforcement learning post-training on LlamaGen-XXL. All samples are generated under identical decoding configurations with CFG scale = 1.0, temperature = 1.0, top-$k$ = 0, and top-$p$ = 1.0.

We also present additional qualitative comparisons on fine-tuning methods like STE~\citep{vqgan} and post-train tokenizer as shown in Fig.~\ref{fig:qual-ste-pt}. 

We further analyze the impact of long-term tokenizer decoder post-training. Surprisingly, although reconstruction loss steadily decreases during training, both FID and IS consistently worsen. This degradation arises because the decoder, trained with teacher-forced ground-truth tokens, gradually learns an overly smooth reconstruction mapping as shown in Fig.~\ref{fig:qual-pt}. It becomes tolerant to token inaccuracies, suppresses high-frequency textures, and specializes in cleaning input tokens that do not match the noisy AR tokens produced during inference. As a result, images retain coarse structure but lose sharpness and perceptual richness.

These observations highlight the inherent limitations of decoder-only fine-tuning: it cannot correct off-manifold token sequences, amplifies the train–inference mismatch, and ultimately smooths away meaningful details. This reinforces the central design choice of our method—fine-tuning the AR generator itself is necessary to align the generated token distribution with the ground-truth image manifold.

Post-training the tokenizer (PT) preserves coarse structure but produces overly smooth textures, while STE-based AR fine-tuning partially improves details yet still suffers from off-manifold token transitions. This is because STE only updates the likelihood of teacher-forced tokens and cannot adjust the AR model’s sampling behavior. In contrast, our VA-$\pi$ optimizes pixel-space rewards for sampled token sequences, yielding sharper textures, more coherent semantics, and overall more realistic generations. These qualitative results highlight that pixel-level reward alignment is essential for correcting sampling errors that STE alone cannot address.

Overall, VA-$\pi$ produces images that are more consistent with class semantics and exhibit improved structural fidelity and perceptual realism compared to the pre-trained LlamaGen-XXL baseline.

\begin{figure*}
    \centering
    \includegraphics[width=0.75\linewidth]{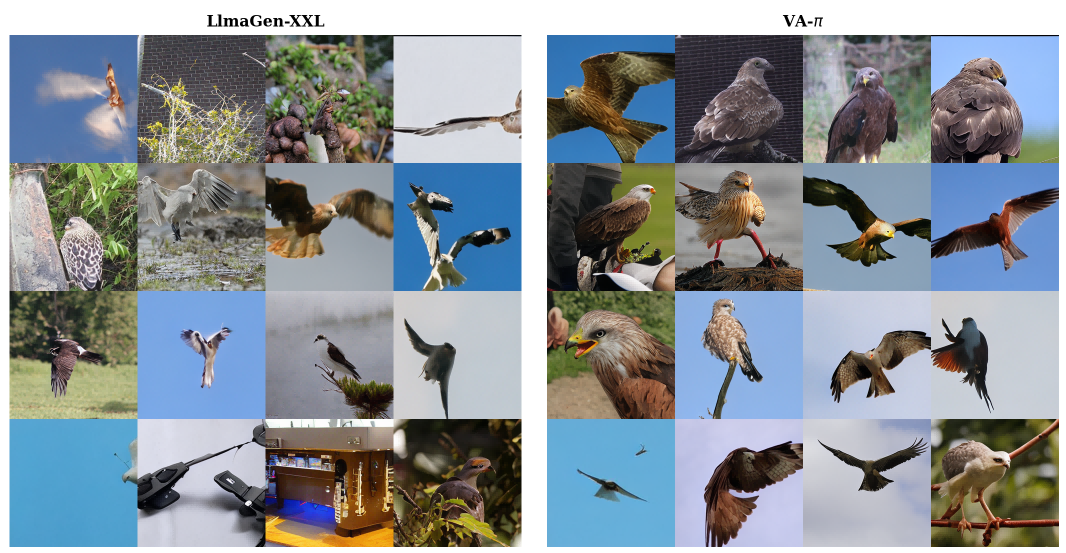}
    \caption{Qualitative comparison on the \texttt{kite} class.}

    \centering
    \includegraphics[width=0.75\linewidth]{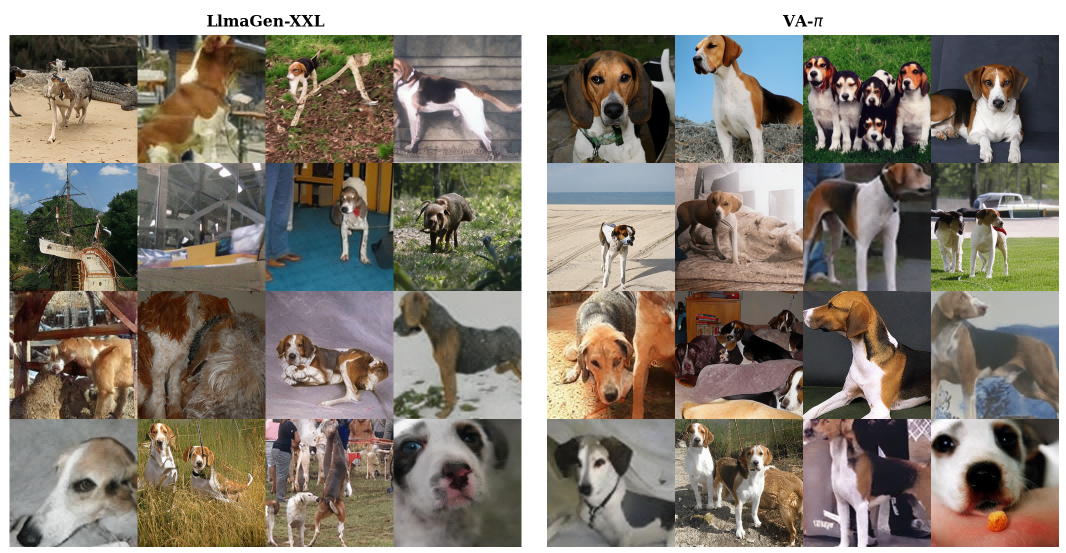}
    \caption{Qualitative comparison on the \texttt{English foxhound} class.}

    \centering
    \includegraphics[width=0.75\linewidth]{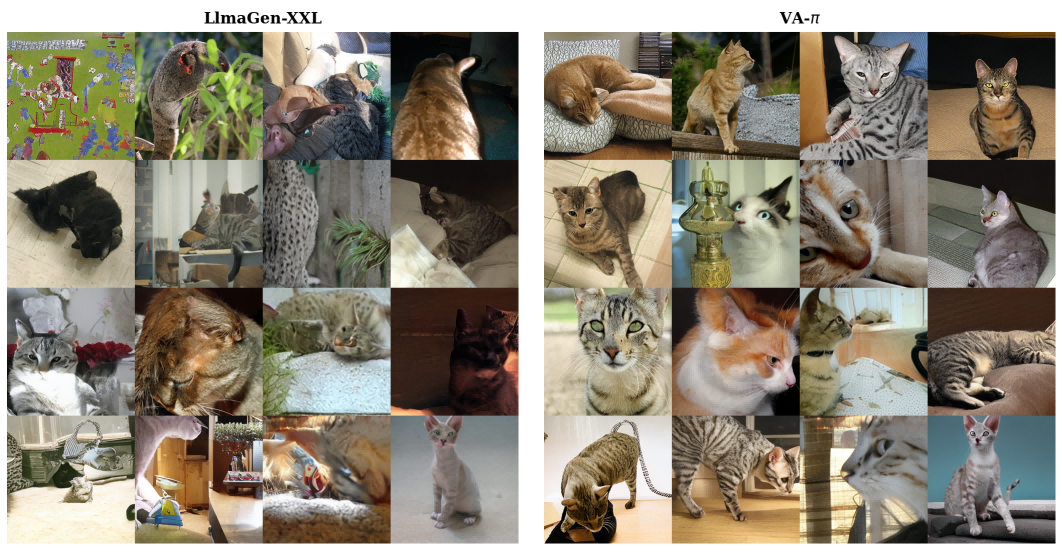}
    \caption{Qualitative comparison on the \texttt{Egyptian cat} class.}
\end{figure*}

\begin{figure*}
    \centering
    \includegraphics[width=0.75\linewidth]{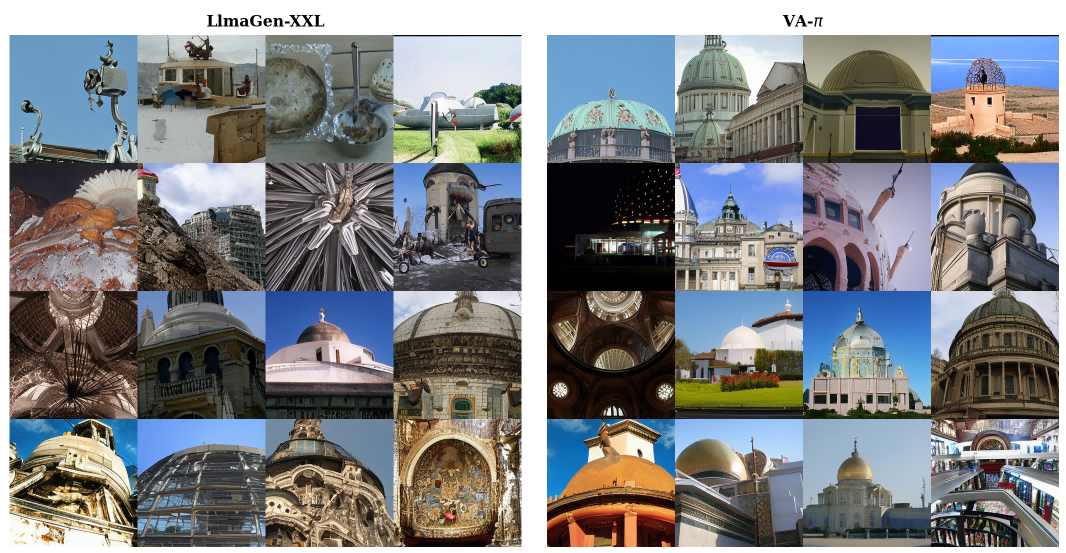}
    \caption{Qualitative comparison on the \texttt{dome} class.}

    \centering
    \includegraphics[width=0.75\linewidth]{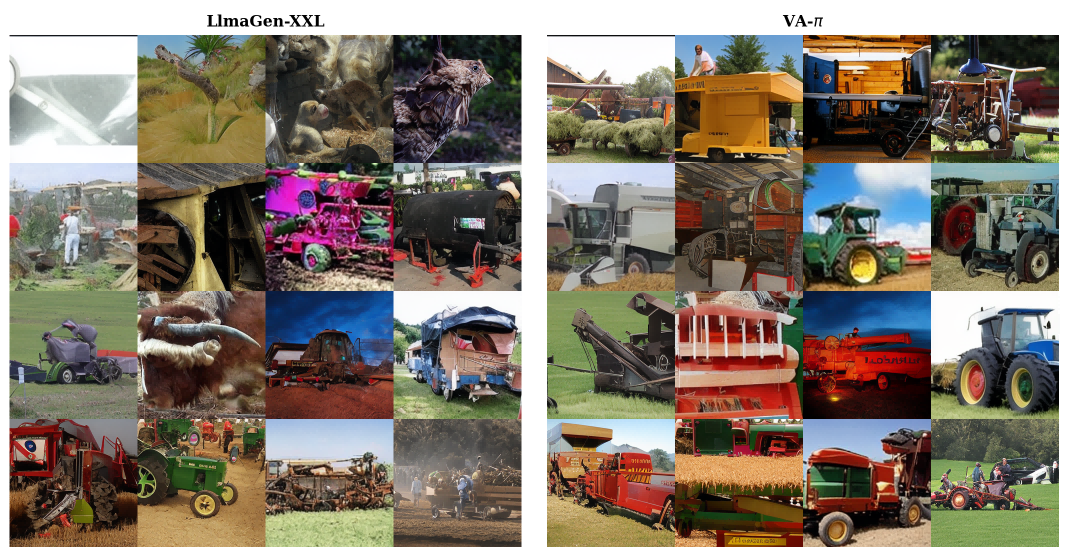}
    \caption{Qualitative comparison on the \texttt{thresher / thrasher / threshing machine} class.}

    \centering
    \includegraphics[width=0.75\linewidth]{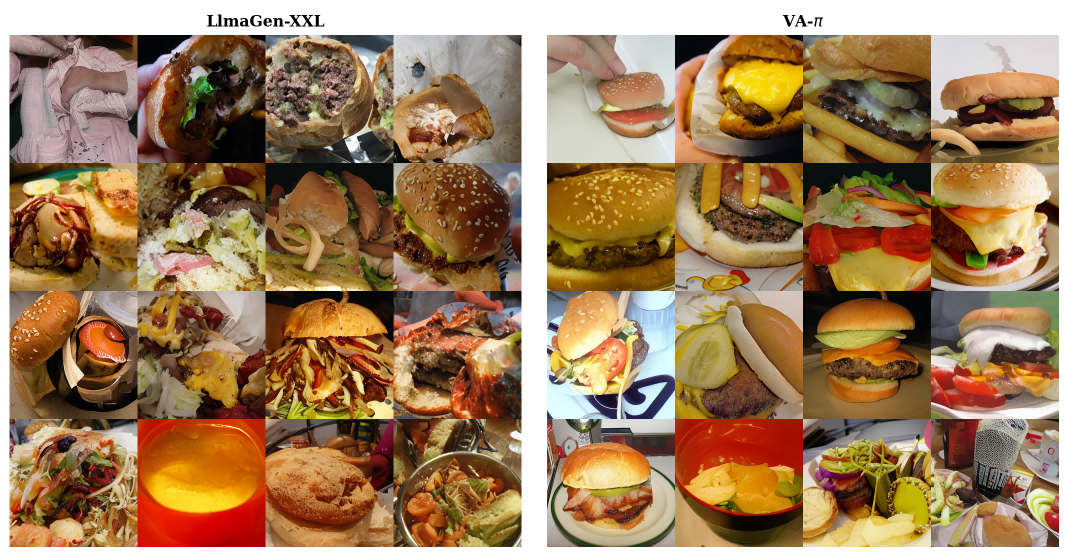}
    \caption{Qualitative comparison on the \texttt{cheese burger} class.}
\end{figure*}

\begin{figure*}
    \centering
    \includegraphics[width=0.75\linewidth]{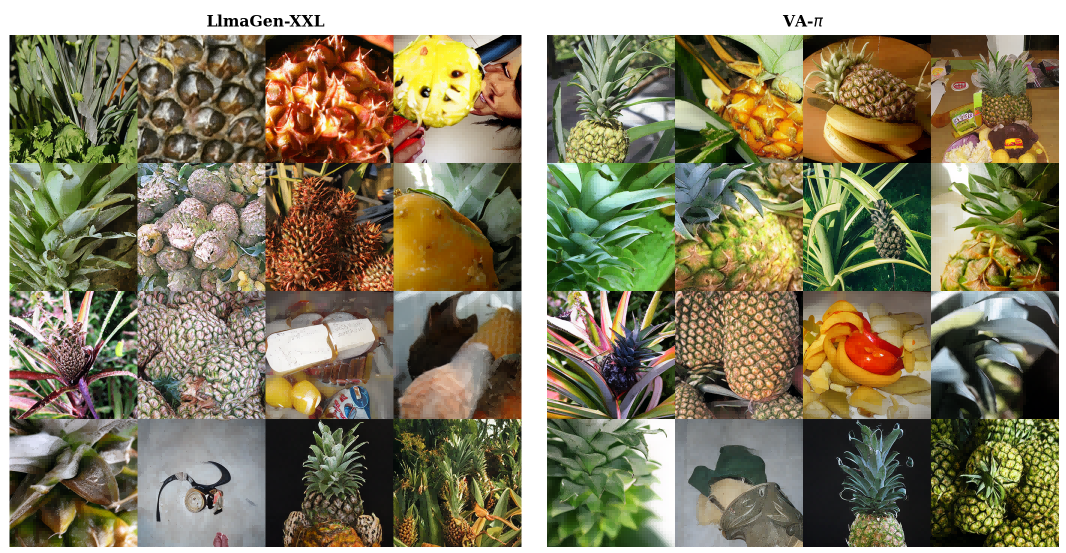}
    \caption{Qualitative comparison on the \texttt{pineapple / ananas} class.}

    \centering
    \includegraphics[width=0.75\linewidth]{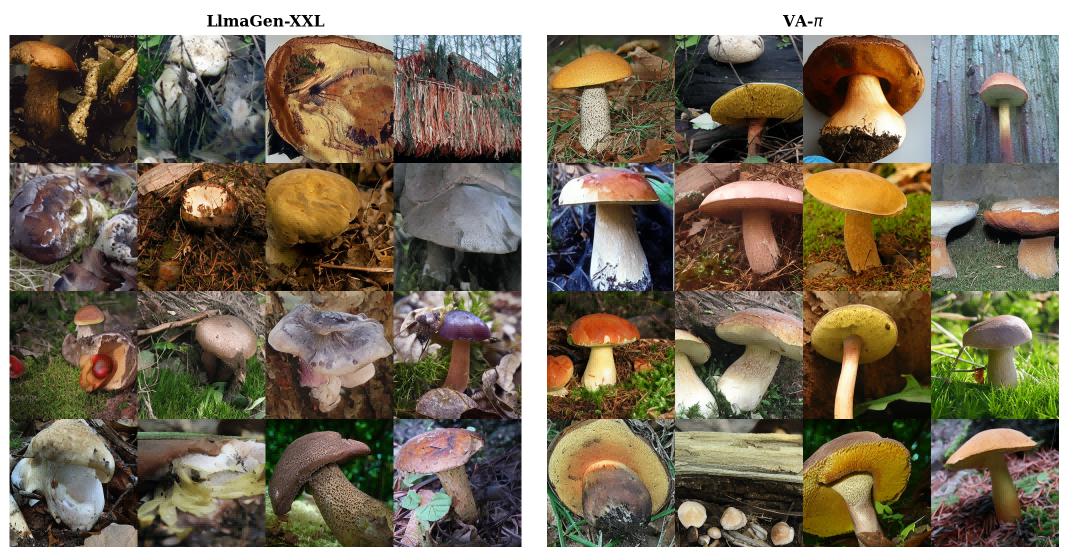}
    \caption{Qualitative comparison on the \texttt{bolete} class.}

    \centering
    \includegraphics[width=0.75\linewidth]{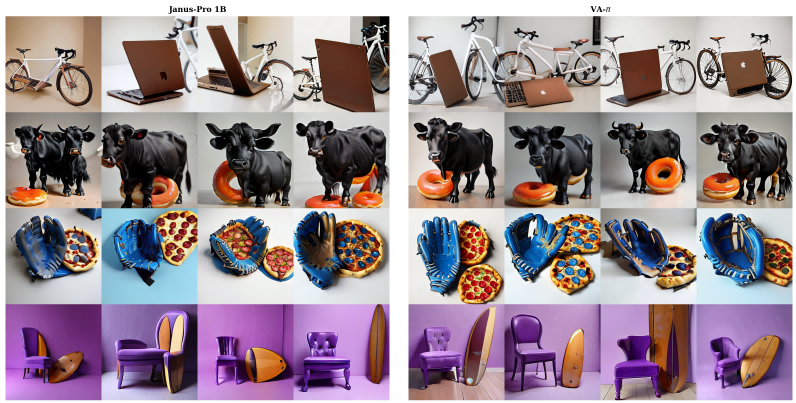}
    \caption{Qualitative comparison on the \texttt{attribute binding} task.}
\end{figure*}

\begin{figure*}
    \centering
    \includegraphics[width=0.75\linewidth]{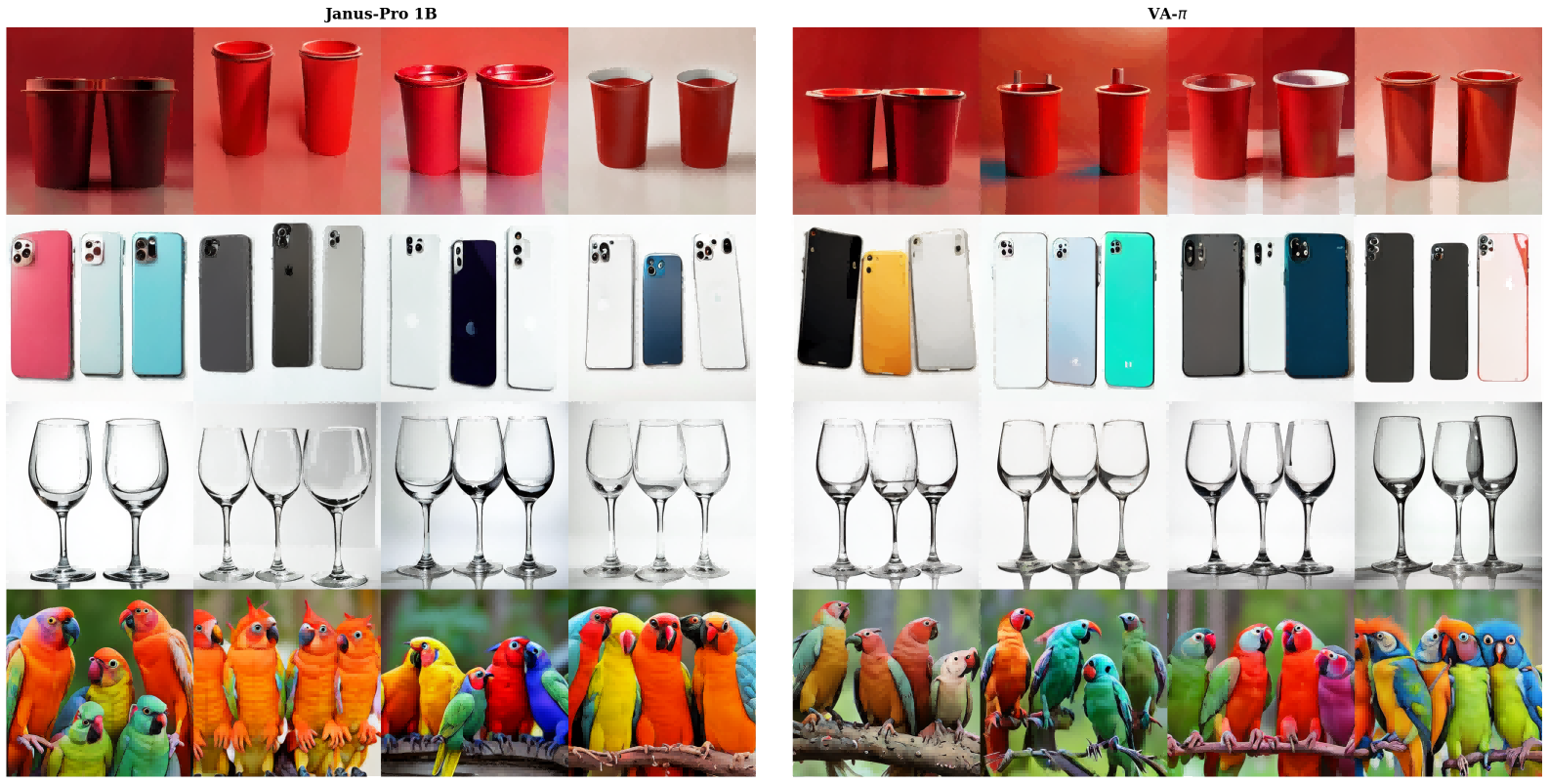}
    \caption{Qualitative comparison on the \texttt{counting} task. }

    \centering
    \includegraphics[width=0.75\linewidth]{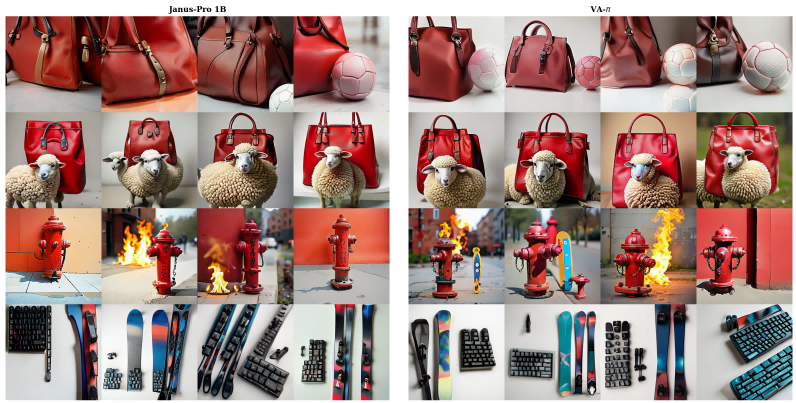}
    \caption{Qualitative comparison on the \texttt{position} task. }

    \centering
    \includegraphics[width=0.75\linewidth]{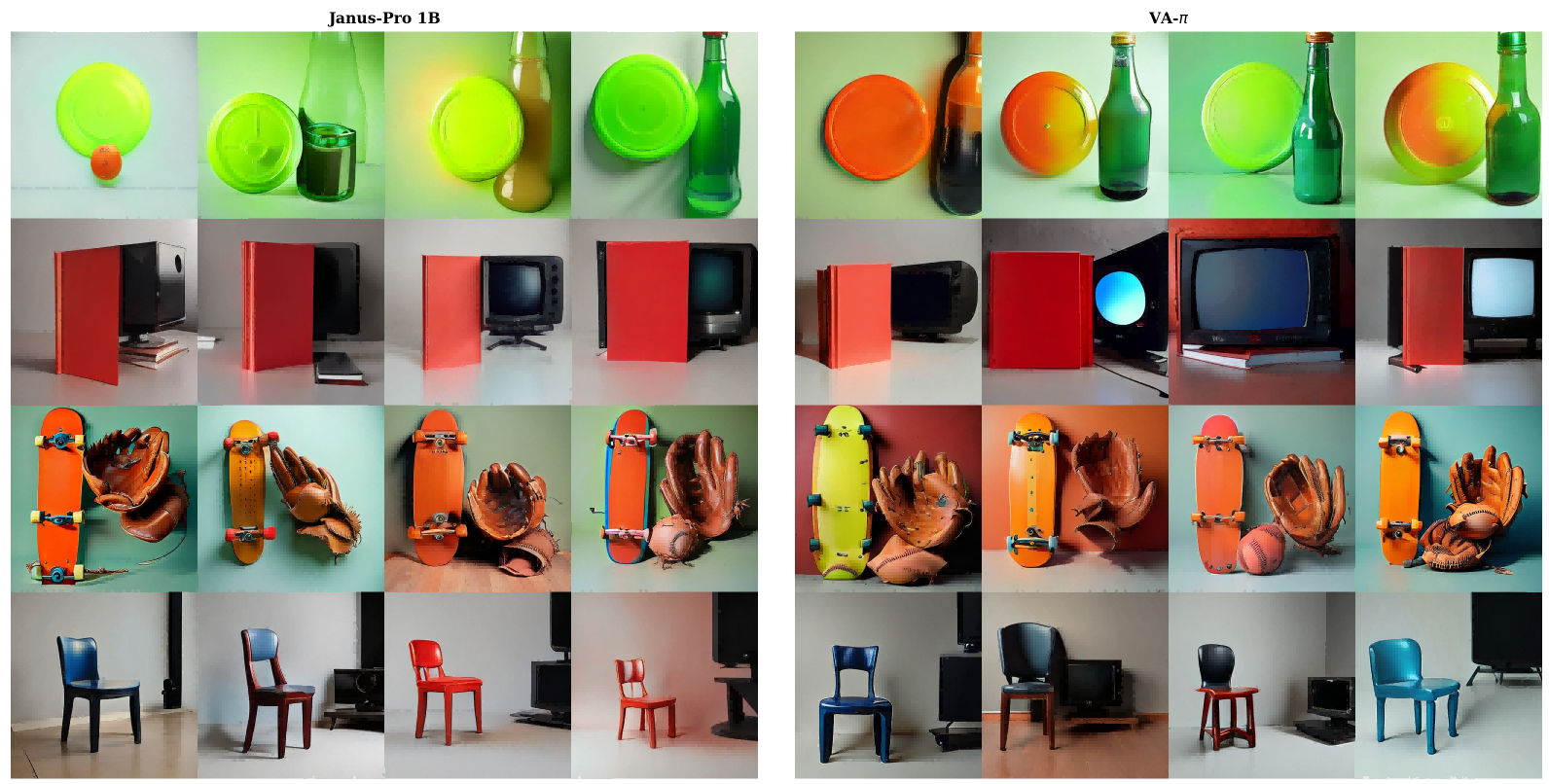}
    \caption{Qualitative comparison on the \texttt{two-object combination} task. }
\end{figure*}

\subsection{Text-to-Image Generation}

We present additional qualitative comparisons on the text-to-image (T2I) generation task, showcasing examples from various prompts from the the GenEval benchmark~\citep{geneval}. Especially, we present images generated from complex tasks: attribute binding, counting, position, and two-object combination. Each comparison visualizes generations from the unified multi-modal baseline Jans-Pro 1B~\citep{januspro} and our VA-$\pi$, which applies reinforcement learning post-training on Jans-Pro 1B. All samples are generated under identical decoding configurations with CFG scale = 5.0, temperature = 1.0, top-$k$ = 0, and top-$p$ = 1.0.
{
    \small
    \bibliographystyle{ieeetr}
    \bibliography{main}
}


\end{document}